\documentclass{article}
\usepackage{amssymb}

\usepackage[preprint]{corl_2025} 
\usepackage[numbers]{natbib}
\usepackage{multicol}
\usepackage{graphicx,subcaption}
\usepackage{amsmath}
\usepackage{enumitem}
\usepackage{adjustbox}
\usepackage{algpseudocodex}
\usepackage{algorithm}
\usepackage{multirow, multicol}
\usepackage{wrapfig,lipsum,booktabs}

\def\Snospace~{\S{}}

\algrenewcommand\algorithmicrequire{\textbf{Input:}}
\algrenewcommand\algorithmicensure{\textbf{Output:}}

\title{Continual Robot Skill and Task Learning via Dialogue}

\author{
  Weiwei Gu\\
  Arizona State University\\
  \texttt{weiweigu@asu.edu}
  \And
  Suresh Kondepudi\\
  Arizona State University\\
  \texttt{nkondepu@asu.edu}
  \AND
  Anmol Gupta\\
  Arizona State University\\
  \texttt{agupt374@asu.edu}
   \And
  Lixiao Huang\\
  Arizona State University\\
  \texttt{lixiao.huang@asu.edu}
   \And
  Nakul Gopalan \\
  Arizona State University\\
  \texttt{ng@asu.edu} \\
}

\begin{document}
\maketitle


\begin{abstract}
Interactive robot learning is a challenging problem as the robot is present with human users who expect the robot to learn novel skills to solve novel tasks perpetually with sample efficiency.  
In this work we present a framework for robots to \emph{continually learn tasks and visuo-motor skills and query for novel skills via dialog} interactions with human users. 
Our robot agent maintains a skill library, and uses an existing LLM to perform grounded dialog interactions to query unknown skills from real human users. 
We developed a novel visual-motor control policy Action Chunking Transformer~\cite{zhao2023learning} with Low Rank Adaptation (ACT-LoRA) that can continually learn novel skills using only a few demonstrations which is critical in human-robot interaction scenarios. The paper has twin goals: Firstly to demonstrate better continual learning in simulation; and secondly, to demonstrate the use of our dialog based learning framework in a realistic human-robot interaction use case.
Our ACT-LoRA policy consistently outperforms a  GMM-LoRA baseline on multiple continual learning simulation benchmarks by achieving $>300\%$ improvements on novel skills, while achieving comparable performance in existing skills.
Moreover, with our IRB approved human-subjects study we demonstrate that our dialog based continual learning framework allows users to teach robots cooking skills successfully (100\%) while spending a higher ratio of time on finishing an auxiliary distraction tasks in the test phase of the study compared to a non-learning language based agent ($p<0.001$).
\end{abstract}


\section{Introduction}

\citeauthor{Chai2019TeachingRN}\citeyear{Chai2019TeachingRN} define natural interaction as an interaction between a human and a robot that resembles the way of natural communication between human beings such as dialogues, gestures, etc. without requiring the human to have prior expertise in robotics. 
 The capability of learning tasks and acquiring new  skills from natural interactions is desirable for robots as they need to perform unique tasks for different users.
One direction of this interaction channel is well studied as instruction following~\cite{ahn2022i, brohan2023rt1, brohan2023rt2}, where the robot performs the tasks requested by the human via natural language. 
Our work focuses on the other side of this communication channel, where the robot starts the conversation with human when it needs help to learn a skill the robot does not know yet. 
Furthermore, our work pushes the boundary of continual learning by introducing a novel visuo-motor policy that continually learns novel skills with as few as three demonstrations.

Human-Robot interaction via language is a well studied problem~\cite{Chai2019TeachingRN, brohan2023rt1, brohan2023rt2, gu2024interactive}. 
Instruction following is the most common approach in task specification and robot execution~\cite{ahn2022i, brohan2023rt1, brohan2023rt2}. 
These methods generally rely on the emergent behaviors of large models, and do not continually learn novel skills. 
To address this issue, some works have proposed continual learning for robot agents~\cite{thrun1995lifelong, lesort2020continual, gu2024interactive, wan2024lotus}. 
There is plenty of recent work that learn visuo-motor skills in a continual setting~\cite{wan2024lotus,xu2023xskill,liu2024tail} as a passive skill discovery problem without user queries.
Other simulation-based approaches introduce robot agents that can understand dialog and take discrete actions~\cite{padmakumar2021teachtaskdrivenembodiedagents}. 
The closest work to us in the human-robot interaction setting is a dialog-based skill teaching pipeline which does not learn end-to-end visuo-motor skills~\cite{grannen2024vocalsandboxcontinuallearning}. 
Moreover, we demonstrate sample efficiency compared to SOTA continual robot learning baseline~\cite{liu2024tail} across multiple simulation benchmarks. 

\begin{figure*}[t]
    \centering
    \includegraphics[width=0.9\textwidth]{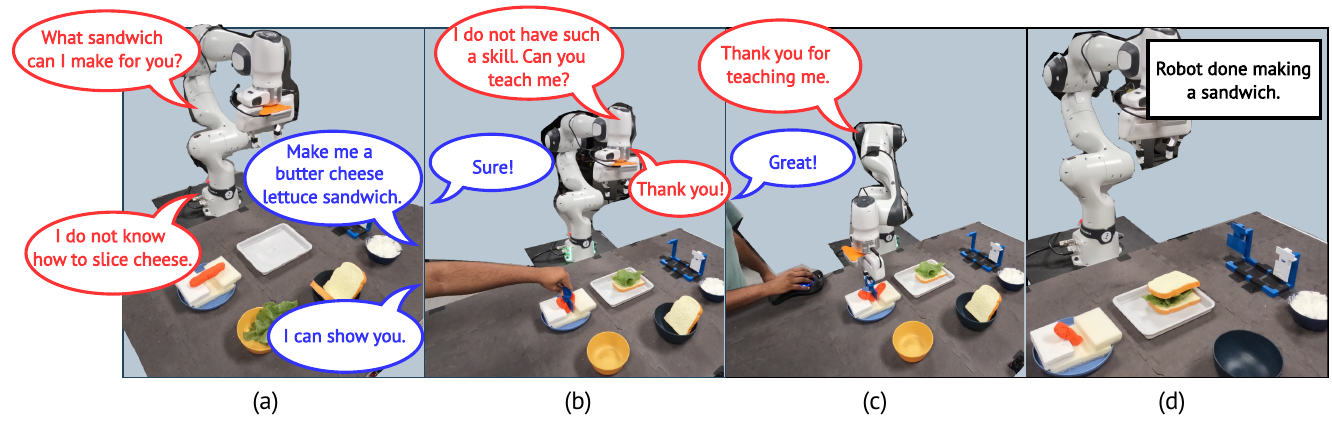}
    \caption{A demonstration of our framework in the user study. (a) The user asks the robot to make a sandwich, some skills to make a sandwich are known but the  dynamic skill to slice cheese is unknown. (b) The human cuts the cheese, by hand, to demonstrate the type of skill needed, but as the robot has never seen such a skill before it asks for tele-operative demonstrations. (c) The user controls the robot to demonstrate said skill $5$ times. (d) The robot learns the novel skill from the human demonstration and is able to complete the entire sandwich on its own in the next interaction.}
    \label{fig:storyboard}
\end{figure*}

We propose a novel framework that learns tasks and novel skills from dialog interactions by querying human users for unknown skills.
To the best of our knowledge, we present the first robot agent that can continually learn dynamic visuo-motor skills in an end-to-end fashion using dialog, which can scale and function in a dynamic visual environment.
Our contributions are as follows:
\begin{enumerate}[leftmargin=5mm, nolistsep]
    \item We present a Continual Learning Aided by Dialog Agent (COLADA) that leverages dialog interactions to keep improving by learning novel visuo-motor skills and task groundings.
    \item We developed a sample efficient Continual Learning algorithm for robots - ACT-LoRA to function in these real world domains. 
    ACT-LoRA achieves an improvements of over $500\%$ on pre-trained skills than the ACT model, and outperforms GMM-LoRA by over $300\%$ on fine-tuned skills across our simulation benchmarks.
    \item Finally, we conduct a within-subjects user study with $16$ users. Our system learns and performs novel skills from non-expert users.  
    COLADA achieved an overall task and novel skill success rate of $87.5\%$, and $100\%$ respectively.
    Furthermore, our framework allows users to spend  a higher ratio of time on finishing other tasks compared to a non-learning language enabled agent with significance ($p<0.001, Z=3.61$) indicating the utility of such learning agents.
\end{enumerate}

\label{sec:intro}




\section{Methods}

\subsection{COLADA}
The goal of our framework is a robot agent that 1) learns high-level plans from dialog interactions with users; 2) queries the user for unknown skills; and 3) learns new skills with only a few instances.
Our robot agent can learn a high level plan from the dialog interactions with the human users. 
We use a language model to map the high-level task specified by the users' utterances into a sequence of low-level motor skills $\tau$.
When needed to perform a motor skill $\tau$, the robot agent first searches for a learned skill using semantic representation, which comes from the language embedding of the linguistic description of the skills. 
This is a challenging question as the robot needs to know what it does not know. 
This work is performed by our queryable skill library.
The robot agent can directly perform the skill $\tau$ whenever it finds a learned skill that aligns with $\tau$ in the semantic space. If $\tau$ is too far in the semantic space COLADA has to learn this skill. COLADA actively tries to learn this novel skill by requesting few robot demonstrations from the human users.
To learn these novel visuo-motor skills from a few robot demonstrations, we developed a novel sample efficient continual skill learning approach ACT-LoRA for this task.
Throughout the interaction with the users, our framework not only learns the novel visuo-motor skill, but also learns to ground language tokens to the skills. This enables our agent to perform the same skill when encountering the same language query at test time.

\begin{figure*}[tb]
    \centering
    \includegraphics[width=0.8\textwidth]{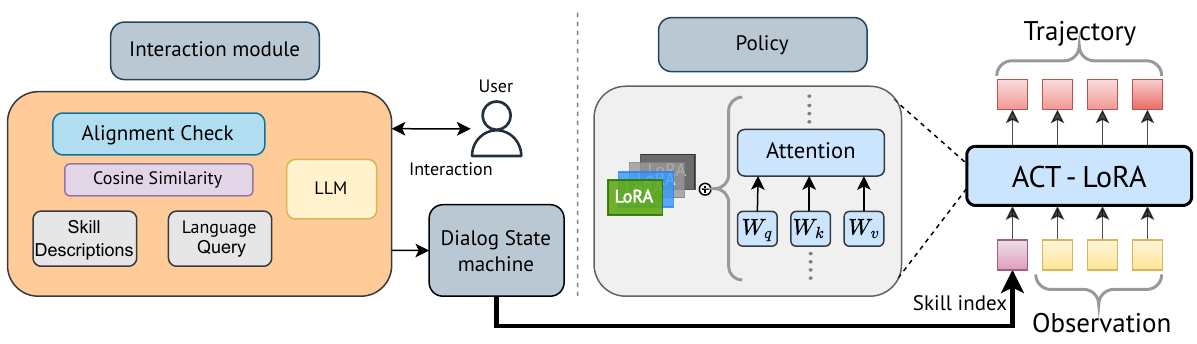}
    \caption{Overview of our COLADA framework. The LLM serves as the interactive module and understands a user's feedback. The skill library provides representations for learned skills and novel demonstrations. The policy model executes the tasks based on the user's instructions. The agent searches for an executable skill by comparing the language representation of the queried skill and language representations of existing skills using a cosine similarity metric. We integrate Low-Rank Adaptor(LoRA) with ACT as our continual learning policy.}
    \label{fig:model}
\end{figure*}
\begin{algorithm}[t]

\caption{The Algorithm for the Dialogue State Machine}
\label{alg:state_machine}
\begin{multicols}{2}
\begin{algorithmic}[1]
\Require{
$\mathcal{O}_0$: Observations; $\mathcal{S} = \{S_1,\dots, S_K\}$: Initial skill library; 
$\pi_\psi, \psi = \{\psi_0, \psi_1, \dots, \psi_K\}$: Policy $\pi$ parameterized by $\psi$,  shared weights $\psi_0$ and skill specific weights $\{\psi_1,\dots, \psi_K\}$; 
$\epsilon_\text{text}$: Skill similarity threshold}
    \State $\mathcal{A} \gets$ GetOrderedListOfSkillsRequested-  \quad ToBeExecuted($\mathcal{O}_0$) 
    \While{$\mathcal{A}$ is not empty}
        \State $\tau \gets \mathcal{A}[0]$ \Comment{pop next skill}
        \State $\mathcal{A} \gets \mathcal{A}[1:]$
        \For{$S_i\in \mathcal{S}$}  \Comment{Compute semantics similarity for each skill}
            \State $s_i \gets$ GetSemanticSimilarity($\tau, S_i$) 
        \EndFor
        \State $j \gets \text{argmax} \{s_1,\dots, s_K\}$ \Comment{Find the skill with the highest score}
        \If{$s_j\geq \epsilon_\text{text}$}
            \State reply $\gets$ ProposeSkillToHuman($S_j$)
            \If{reply$=$agree}
                \State ExecuteSkill($S_j, \pi_\psi$)
                \State Continue \Comment{skip line 13-18}
            \EndIf
        \EndIf
        \State $r\gets$AskForRobotDemonstration($\tau$)
        \State $\psi_{K+1}\gets$FinetunePolicyForNewSkill($r$)
        \State $S_{K+1} \gets \tau$
        \State $\psi = \psi \cup \{\psi_{K+1}\}$
        \State $\mathcal{S} \gets \mathcal{S} \cup \{S_{K+1}\}$ 
        \State $K\gets K+1$\Comment{Add $\tau$ to the skill library}
            
        
    \EndWhile
\end{algorithmic}
\end{multicols}
\end{algorithm}

\subsection{Interaction Module using a Large Language Model (LLM)}
Our robot agent maintains a dialog state $s^d$ using an internal state machine. 
We describe the state machine in Algorithm~\ref{alg:state_machine}. 
The dialog state machine maintains a skill library $\mathcal{S}=\{S_1,\dots, S_K\}$, where $S_1, \dots S_K$ are the skills that are known to the agent. 
The interaction module first picks the sequence of skills to perform $\mathcal{A}$ from an initial dialog with the users.
The agent then tries to perform each skill $\tau$ in the sequence $\mathcal{A}$ in order.
When encountering a skill $\tau$, the agent first computes a semantic similarity score $s_i$ between $\tau$ and each skill in the skill library $S_i$ using the CLIP embedding~\cite{radford2021learning} of their linguistic descriptions.
The robot can directly perform a matching skill $\tau$ if it finds a known skill $S_j$ with a similarity score ($s_j$) higher than a pre-defined threshold $\epsilon_\text{text} = 0.95$. This threshold was determined by testing during the pilots.
If no matching skill is found, the interaction module explains the confusion of the agent using dialog, and actively tries to learn this novel skill by requesting few robot demonstrations from human users.
This interaction module is given the autonomy to continue the dialogue with the user until it acquires the designated information for the agent in any round of the interaction. Further details can be found in Appendix~\ref{app:recreation}.


\subsection{ACT-LoRA as Visual-motor Policy}

\textbf{Adapters and Visuo-motor policies.} 
\citet{liu2024tail} extended Low Rank Adapter(LoRA) into robotics and demonstrated that adapter-based methods can enable a simulated robot agent to continually learn new skills without forgetting the old ones with the TAIL model.
However, the TAIL model heavily relies on pre-training with large scale data, and hence is not data efficient.
On the other hand, Action Chunking Transformer(ACT) demonstrated robust performance on fine-grained tasks even with limited data~\cite{zhao2023learning}, but cannot be directly used for continual learning due to catastrophic forgetting.
\emph{Our insight is that augmenting the ACT architecture with LoRA weights can allow the model to continually learn new skills without catastrophic forgetting, while still retaining the fine-grained control capability and data-efficiency of the ACT architecture.} We demonstrate this through extensive simulation experiments.

\textbf{Continual Imitation Learning.} 
The robot agent is equipped with a continually learning policy $\pi_\psi$ which was previously trained on a set of skills $\mathcal{S}=\{S_1, \dots, S_K\}$.
The policy $\pi_\psi$ is parameterized by $\psi=\{\psi_0,\dots,\psi_K\}$, where $\psi_0$ is the weights for the basic ACT architecture that are shared by all skills, and $\psi_1,\dots, \psi_K$ are weights for the low-rank adapters that are specific for each skill. 
Our policy can be optimized with a behavior cloning $L_1$ loss for action chunks similar to previous work~\citep{zhao2023learning}. This loss function can be written as:
$\mathcal{L}=||\pi(\hat{a}_{t:t+k}|o_t, z;\psi) - a^*_{t:t+k}||,$
where $\pi(\hat{a}_{t:t+k}|o_t, z;\psi)$ is the predicted action chunk from the policy and $a^*_{t:t+k}$ is the ground truth action chunk. 
For each skill $S_i$, the policy $\pi$ predicts the action chunks for this skill using the shared base weights of the model $\psi_0$ and adapter weights $\psi_i$ for this skill. Therefore, for skill $S_i$, the loss $\mathcal{L}$ can be written as:
$\mathcal{L} = ||\pi(\hat{a}_{t:t+k}|o_t, z;\psi_0, \psi_i) - a^*_{t:t+k}||$.
During pre-training, both the model's parameter $\psi_0$ and the skill specific adapter weights $\psi_i$ are active, and receive gradient updates from this loss function.

When learning a new skill $S_{K+1}$, we freeze the shared weights $\psi_0$ and the adapter weights for other skills $\psi_1,\dots, \psi_K$, and only the adapter weights for the newly learned skill $\psi_{K+1}$ receive the gradient updates from the loss function $\mathcal{L} = ||\pi(\hat{a}_{t:t+k}|o_t, z, \psi_{0}; \psi_{K+1}) - a^*_{t:t+k}||$. 
Under this fine-tuning strategy, the policy only updates the adapter weights for the newly introduced task, and any weights that are already learned are not changed during fine-tuning. As a result, the model can learn the novel skill without affecting the performance of any known skill.
Similar to previous work~\cite{liu2024tail}, we found LoRA to be more effective in continual learning than other parameter-efficient fine-tuning(PEFT) methods such as prefix tuning~\cite{li2021prefixtuningoptimizingcontinuousprompts} in our preliminary experiments.


\subsection{Human-Subjects Experiment}
We now look to use our COLADA framework on an IRB approved human subjects experiment. 

\textbf{Making Sandwiches with User Inputs on Fraka FR3 Robot.}
We chose a sandwich making as the domain for user study as it has dynamic tasks requiring contact.
Moreover, sandwich-making is a multi-step process, allowing the robot agent and the participants to have multiple rounds of interactions.
Specifically, the robot helps construct two different sandwiches, including a veggie sandwich and a lettuce sandwich with butter, where users need to teach the robot to pour pepper and apply butter respectively as the robot lacks the full suite of skills to complete the tasks specified by the user. 
When a skill is missing the human user can teach these skills to the robot using a teleoperation spacemouse. The users speak to the robot and we use transcribed inputs to our LLM based dialog state machine.
The exhaustive details of the robotics setup can be found in Appendix~\ref{app:detailed_study}.
Throughout the system's operation, picking and placing robot tools is done by pre-specified waypoints because grasping a tool is not our focus. Figure~\ref{fig:storyboard} demonstrates a scenario in our sandwich making domain. 

\textbf{Baselines.} 
Our objectives in the user study is twofold: Firstly, we want to understand the use of language to ask for help. Secondly, we want to quantify if interactive learning is actually efficient to the user especially because robot teaching is a burden on the user. 
We compared against two baselines -- An \emph{inarticulate agent} that keeps solving a task even though some skills for a task might be unknown. This is similar to an agent that cannot reason about known vs unknown skills. Secondly, an \emph{inverse semantics agent} that knows which skills it does not know but asks for human help every time it reaches an unknown skill, which is similar to existing work~\cite{Tellex2014AskingFH}. 
All three agents use ACT-LoRA as the robot policy. 
We did not compare against ACT or GMM-LoRA in the human-subjects study. 
This is because ACT is not a continual learning policy and is destined to fail in the test phase of the study.  
GMM-LoRAs fine-tuned skill performance is order of magnitudes worse compared to ACT-LoRA in the few ($5$) demonstrations setting. When tested during pilots GMM-LoRA crashed the robot into the table twice, for which we considered unsafe for a further human-subjects exposure.
We include more detailed discussions in Appendix~\ref{app:detailed_study}.


\textbf{Study Design and Measures.} We designed a novel IRB approved two-phase study. We collected data for skill learning in the first phase and then ran a learned policy on the agent in the second phase. 
During the first phase - \emph{the interaction phase}, participants interact with the robot to specify a sandwich type and teach the robot novel skills as needed.  In the second phase - \emph{the evaluation phase,} participants request the robot to make the same sandwich and evaluate its performance. 
In each phase the participants were expected to write emails as an auxiliary task.
The two sessions are at least one day apart, allowing $5\texttt{+}$ hours of time to train ACT-LoRA using user data. This makes our study two separate $1 \times 3$ within-subjects experiment to measure our framework's ability to perform known skills and tasks, while continually learning novel skills from non-expert human users.

For objective metrics we measured the overall success rate (SR) of completing the entire sandwich and the success rate for completing each independent skills. We also measured time spent teaching the robot and interacting with the robot in each phase. Number of emails and words written for the auxiliary tasks were also measured along with the subjective metrics of  Godspeed Likability sub-scale~\cite{godspeed}, System Usability Scale (SUS)\cite{sus}, and the NASA TLX~\cite{tlx}. These post-experiment measures were taken for each phase of the study.
  

\begin{table*}[t]
\centering
\resizebox{\linewidth}{!}{%
\begin{tabular}{c|ccccc}
\toprule
Model    & Pre-trained Skills(1000 traj.)   & Fine-tune Skills(1000 traj.) & Overall Success Rate(1000 traj.) & Fine-tune Skills(5 traj.) & Overall Success Rate(5 traj.)\\ 
\midrule
ACT-LoRA& $\mathbf{60.75 \pm 2.40}$ & $54.00 \pm 9.73 ^ *$  & $\mathbf{59.40 \pm 1.52}$ & $77.67 \pm 9.36$ & $\mathbf{64.13 \pm 1.80}$  \\
GMM-LoRA& $26.08 \pm 4.02$ &  $13.33 \pm 4.50$ & $23.53 \pm 2.99$ & $16.67 \pm 4.92$& $24.20 \pm 3.72$\\
ACT& $9.25 \pm 2.51$ & $62.00 \pm 8.84 ^ *$   & $19.80 \pm 1.69$ &   $\mathbf{95.00 \pm 4.22}$ & $26.40 \pm 2.45$ \\

\bottomrule
\end{tabular}%
}
\caption{Experimental results on RLBench dataset. $*$ indicating  similar best performance.
}


\label{tab:rlbench}
\end{table*}
\textbf{Procedure.}
Participants first filled out the consent form and a pre-study survey. 
Then, we handed out a general introduction of the experiment and administered the two phases sequentially. 
Before each phase, a demonstration video and the instructions for the corresponding phase were provided to the participant. 
The instruction manual and videos are provided in the supplementary.

\emph{The interaction phase:-} Here the participant requested the robot agent to make one of two sandwiches. During the process, all methods asked the participant for task knowledge  or robot demonstrations using dialog. The participant answered task knowledge-relevant questions and provided robot demonstrations on request. 
As an auxiliary task the participants also wrote emails for various tasks on a computer while the robot made the sandwich. 
At the end of the interaction phase, the participant was asked to fill out a survey to evaluate the subjective experience with the robot. The participants observed all the baselines of COLADA, Inverse semantics and Inarticulate agents in random order.

\emph{The evaluation phase:-} The participant returns after skill learning and asks the robot agent to make the exact same sandwich as requested in the interaction phase. The participant again performs auxiliary tasks while the robot finishes the sandwich. In some baselines the participant might still need to help the robot to perform unknown skills.  Finally they fill a post-survey to evaluate the robot's subjective performance for all baselines. 

\label{sec:methods}


\section{Experimental Results}

In this section, we present two sets of experimental results.
Firstly, we present the results on continual imitation learning in the simulated RLBench environment~\cite{james2019rlbench} and on three suites of the LIBERO environment.
We then present the results for the human subjects study, which demonstrates that our framework can learn visuo-motor skills by interacting with real human users.

\subsection{Simulation Experiments on Continual Imitation Learning}
We chose two visuo-motor policies, ACT~\cite{zhao2023learning} and GMM-LoRA, as baselines to compare against our model on continual learning. 
ACT~\cite{zhao2023learning} is a SoTA visuo-motor policy that is able to perform dynamic tasks that require high precision, and GMM-LoRA is our re-implementation of the SoTA continual learning policy TAIL with the help of the original authors~\cite{liu2024tail}. 
We use the name GMM-LoRA instead of TAIL because we did not have the large amount of robot data required to pre-train the full GPT style TAIL model. TAIL's results were demonstrated only in simulation~\cite{liu2024tail} and we are demonstrating the challenges in putting these continual learning frameworks on a real robot. 

We report success rate on both RLBench and LIBERO simulators for $20$ rollouts. In both domains, all  three models go through a pre-train, fine-tune training schema.
In Tables~\ref{tab:rlbench} and~\ref{tab:libero}, column \textbf{Pre-trained skills($x$ traj.)} measures the policies' average success rate on the skills that policies are pre-trained on after fine-tuning on $x$ demonstration trajectories. Columns \textbf{Fine-tuned skills($x$ traj.)} and \textbf{Overall Success Rate($x$ traj.)} measure the policies' average success rate on the new skills, and the average success rate for both the pre-trained and fine-tuned skills respectively. 
We use a static initial configurations when evaluating models fine-tuned with $5$ trajectories for RLBench as the performance is close to zero across the board if the objects can move in this extreme case.

We first present our experiments on RLBench~\cite{james2019rlbench}. 
A total of $15$ skills are chosen from the pre-defined skills of the environment. 
We separate these skills into $5$ different splits, and perform a five-split validation in Table~\ref{tab:rlbench}.
ACT-LoRA consistently performs well in both pre-trained and fine-tuned skills, which also explains why ACT-LoRA has the highest overall success rates across the board as demonstrated in Table~\ref{tab:rlbench}.  
 ACT-LoRA outperforms GMM-LoRA on fine-tune skills by $435.5\%$ on average demonstrating its effectiveness in learning novel skills. Our results demonstrate that ACT-LoRA is a better continual learning policy as it remembers pre-trained skills after fine-tuning better, and while also learning the novel skills better than existing baselines.  

We also conduct simulation experiments on three suites of the LIBERO dataset~\cite{liu2023liberobenchmarkingknowledgetransfer}, the spatial suite, the object suite, and the goal suite.
Similarly, for each task suite we split the $10$ skills into $5$ different splits of skills and perform a five-split validation.
We report the experiment results and statistics from the five-split validation in Table~\ref{tab:libero}.
Although GMM-LoRA out-performs ACT-LoRA at pre-trained skills in the LIBERO-Object task suite by a small margin, ACT-LoRA consistently out-performs GMM-LoRA on all other pre-trained or fine-tuned skills across all the three task suites. ACT-LoRA outperforms GMM-LoRA on fine-tune skills by $493.6\%$ on average across the three suites of LIBERO.
As ACT is not a continual learning pipeline it consistantly forgets pre-trained skills, and we consistantly demonstrate order of magnitude improvements on GMM-LoRA over fine-tuned skills. 
These demonstrate that ACT-LoRA has the overall most stable performance across all the three models and was chosen by us to be used in our user studies.

\begin{table*}[tbh]
\centering
\resizebox{\linewidth}{!}{%
\begin{tabular}{cccccc}
\toprule

Model    & Pre-trained Skills(50 traj.)   & Fine-tune Skills(50 traj.) & Overall Success Rate(50 traj.) & Fine-tune Skills(5 traj.) & Overall Success Rate(5 traj.)\\ 
\midrule
 & \multicolumn{5}{c}{\textbf{LIBERO-Spatial}}\\
 \cmidrule(lr){2-6}
ACT-LoRA& $65.38 \pm 4.51 ^ *$ & $40.50 \pm 6.09$ & $\mathbf{60.40 \pm 4.20}$ & $35.50 \pm 8.27$  & $59.40\pm 4.40 ^ *$\\
GMM-LoRA& $64.75 \pm 2.49 ^ *$ &  $9.00 \pm 5.16$& $53.60 \pm 1.70$ & $6.00 \pm 2.92$& $53.0 \pm 2.21 ^ *$\\
ACT& $0.03 \pm 0.02$ & $\mathbf{68.50 \pm 6.50}$ & $13.90 \pm 1.31$ & $\mathbf{55.00 \pm 7.66}$ & $11.20 \pm 1.43$\\

\midrule
 & \multicolumn{5}{c}{\textbf{LIBERO-Object}}\\
 \cmidrule(lr){2-6}
ACT-LoRA& $67.00 \pm 2.20$ & $68.00 \pm 8.57 ^ *$ & $67.20 \pm 1.50 ^ *$ & $48.00 \pm 10.23 ^ *$  & $63.20 \pm 1.60 ^ *$ \\
GMM-LoRA& $\mathbf{77.75 \pm 1.90}$ &  $15.00 \pm 5.65$ & $65.20 \pm 2.15 ^ *$& $14.00 \pm 5.89$ & $65.00 \pm 1.08 ^ *$\\
ACT& $12.88 \pm 2.78$ & $63.00 \pm 9.33 ^ *$ & $22.90 \pm 2.45$ & $35.50 \pm 7.92 ^ *$ & $17.40 \pm 3.45$\\

\midrule
 & \multicolumn{5}{c}{\textbf{LIBERO-Goal}}\\
 \cmidrule(lr){2-6}
ACT-LoRA& $73.63 \pm 2.96 ^ *$ & $\mathbf{49.00 \pm 8.54}$ & $\mathbf{68.70 \pm 3.70}$ & $23.00 \pm 8.57 ^ *$  & $63.50 \pm 4.00 ^ *$ \\
GMM-LoRA& $75.38 \pm 1.63 ^ *$ &  $10.50 \pm 5.61$ & $62.40 \pm 1.39$ & $3.5 \pm 2.92$ & $61.00 \pm 1.72 ^ *$\\
ACT& $0.00 \pm 0.00$ & $19.50 \pm 3.66$ & $3.90 \pm 0.73$ & $10.50 \pm 4.57 ^ *$ &  $2.10 \pm 0.91$\\

\bottomrule
\end{tabular}%
}
\caption{Results on three suites of LIBERO dataset. $*$ indicating  similar best performance.} 

\label{tab:libero}
\end{table*}

\subsection{Human Subject Study Results on a Franka FR3 Robot}

We conducted an IRB approved study with $16$ participants (with age ranges $23.44\pm 0.51$) and $20$ pilot subjects. 
We had $8$ female subjects ($50.00\%$ of the user study). 
Our participants have $0.5 \pm 0.32$ and  $4.69 \pm 0.66$ years of experience in robotics, and   computer science respectively.
The subjects spent $120$ minutes and $75$ minutes in the interaction and evaluation phases respectively. They were compensated with a $\$35$ Amazon gift card.  
We present our objective success rate results in Table~\ref{tab:study}, and objective metrics on the distraction tasks in Table~\ref{tab:interface_time_ratio}. 
We test normality for each metric with Shapiro-Wilk test. If the data from such metric passes the normality test($p>0.05$), we apply a parametric statistical test. Otherwise, we report the results of a non-parametric statistical test. The detailed results of the statistics tests can be found in Appendix~\ref{app:detailed_study} and \href{https://sites.google.com/view/corl-25-dialog/home}{the associated webpage}.

\textit{\textbf{RQ1:} Can COLADA learn novel skills from and finish tasks for novice human users?} We find that COLADA is able to learn novel visuo-motor skills from novice human users with just $3$ tele-operation demonstrations. 
As shown in Table~\ref{tab:study}, our ACT-LoRA policy learns the novel skills with the success rate of ($100\%$) and performs existing skills with the success rate of $88.89\%$. 
Additionally, our COLADA achieves $93.75\%$ and $81.25\%$ sandwich completion rate in the two phases of the study respectively as shown in Table~\ref{tab:study}. This is comparable to the performance of the inverse semantics agent, which relies on help from the users and cannot finish the task independently.

\textit{\textbf{RQ2: }Is it actually efficient for users to teach our robot novel skills?} 
We expect subjects to spend more time teaching COLADA in the first phase and then using almost no time helping the agent in the second phase. 
The ratio of the time that participants spend on the distraction email writing task to the total time that they spend with the agent is presented in Table~\ref{tab:interface_time_ratio}.
More importantly, we also find that COLADA allows participants to spend a higher ratio of time writing emails than both the inverse semantics agent ($p<0.001, Z=3.61$) and the inarticulate agent($p<0.001, Z=4.17$) in phase two with significance, as indicated by Wilcoxon Signed-Rank test.
These results suggest that the ability of learning enables the agent to complete the tasks autonomously, and eventually improve the time efficiency for users.


\textit{\textbf{RQ3:} Is COLADA preferred over the inverse semantics agent in subjective metrics?} 
We found that participants rate COLADA to be more usable with SUS compared to the inverse semantics agent($p=0.044, t=1.83$). 
For other subjective metrics, we failed to find any significance between COLADA and the inverse semantics agent. Details and discussions can be found in Appendix~\ref{app:detailed_study}.

Overall, we notice users being able to teach visuo-motor tasks few-shot to the robot and having a higher percentage of time to their auxiliary tasks such as writing emails using COLADA. We provide a demonstration video as a supplement to show how COLADA learns tasks and skills from its first interaction with a user, and performs the tasks fully autonomously in the second interaction.

\begin{table}[tbh]
        \centering
        \resizebox{\textwidth}{!}{%
        \begin{tabular}{c|cc|ccc}
            \toprule
            Agent    & \multicolumn{2}{c|}{Phase 1}       & \multicolumn{3}{c}{Phase 2}   \\ 
             & Sandwich SR & Pre-train SR & Sandwich SR  & Few-shot SR & Pre-train SR\\
             \cmidrule(lr){1-1}\cmidrule(lr){2-3} \cmidrule(lr){4-6}
            COLADA& $93.75\%(15/16)$& $97.92\%(47/48)$ & $81.25\%(13/16)$& $100.00\%(16/16)$ & $91.67\%(44/48)$\\
            Inverse Semantics& $81.25\%(13/16)$ & $93.75\%(45/48)$& $87.50\%(14/16)$ & N/A & $91.67\%(44/48)$\\
            Inarticulate& $0.00\%(0/16)$ & $93.75\%(15/16)$& $0.00\%(0/16)$ & $0.00\%(0/16)$  & $87.50\%(14/16)$ \\
            \bottomrule
        \end{tabular}%
        }
        
        \caption{Objective metrics of the three agents in the human-subjects study.
        \textbf{Sandwich SR} is the overall task success rate; \textbf{Pre-train SR} and \textbf{Few-shot SR} are the success rates of the pre-train skills and the newly learned skills in phase 2 respectively. \textbf{Few-shot SR} for inverse semantics agent is not applicable as it always asks for help.}
        \label{tab:study}
\end{table}

\label{sec:result}

\section{Discussion}

In our analysis, we have demonstrated that COLADA can continually learn skills from dialog interactions with novice human users with our end-to-end neural network policy ACT-LoRA, and complete the requested tasks automatically with a success rate comparable to an upper bound of a human aided inverse semantics agent. 
This showcases that ACT-LoRA model is robust as a continual learning policy under low data regimes, whereas other SoTA continual learning policies such as TAIL~\cite{liu2024tail} rely on fine-tuning large vision backbone on environment data and also require large scale of data for training, which is not accessible  most of the time in real-world robot applications. Additionally, our ACT-LoRA model also outperforms GMM-LoRA, a scaled down version of the TAIL model with similar number of parameters to ACT-LoRA, on fine-tuned skills in simulation experiments. This further demonstrates that ACT-LoRA is more suitable for continual skill learning. 
Additionally, we have also demonstrated that our end-to-end continual learning method can learn contact rich  dynamic skills from novice users, such as applying butter and pouring pepper. We believe such methods have more potential to scale compared to existing DMP and keypoint based approaches~\cite{grannen2024vocalsandboxcontinuallearning}. 

\label{sec:discussion}

\section{Related Work}

\textbf{Skill Discovery and Continual Learning.}
The area of visuo-motor continual learning attracts a lot of attention~\cite{wan2024lotus,xu2023xskill,liu2024tail}.
Previous work discovers new skills for robots from robot trajectories~\cite{wan2024lotus}, even from human demonstrations~\cite{xu2023xskill}.
Another line of work~\cite{liu2024tail, liang2022transformer} extended low-rank adaptation techniques~\cite{hu2021lora} to robotics and developed robot policies that can continually learn new skills without catastrophic forgetting. 
However, these frameworks assume the presence of the demonstrations for the new skills, and only discover skills in a passive fashion. 
This work is a more natural setup for a language enabled continual learning agent in the real world. 
Furthermore, our agent requires only three demonstrations from the user to learn the new skill, which improves the data efficiency over existing visuo-motor continual learning methods~\citep{liu2024tail, wan2024lotus}.


\textbf{Human-Robot Dialogue.} 
Human-Robot dialog is a mature problem~\cite{dai2024think, Tellex2014AskingFH, Thomason2020JointlyIP, ijcai2018p1, padmakumar2021teachtaskdrivenembodiedagents}. 
Grounded with perceptive inputs from the environment, LLMs have been used in robotics research generate executable plans~\cite{ahn2022i}.
Furthermore, other work use LLMs to ask for human feedback for the robot agents demonstrating the importance of dialog~\citep{ren2023robots, dai2024think}.
However, we are attempting to learn continuous visuo-motor skills on the robot by asking for help and not specify plans. 
Recently, \citet{grannen2024vocalsandboxcontinuallearning} demonstrated a dialog-based skill learning approach, however they are either learning static visuo-motor pick and place tasks or learning dynamic skill with keypoints and dynamic movement primitives (DMPs). We do not make any such assumptions and continually learn end-to-end visuo-motor skills from user data in a few-shot setting allowing for better generalizability. 

\textbf{Active Learning.} Our work is also related to active learning, where a learning agent actively learns new skills by asking for demonstrations~\cite{Thomason2020JointlyIP, pmlr-v78-maeda17a,Chernova_2009, chernova_2007}.
Defining an appropriate metric that triggers the request for assistance becomes the key research problem in this domain.
Previous work~\cite{Thomason2020JointlyIP} measures the semantic similarity between a new concept and the known concepts to ask for classifier labels.
Other work~\citep{Chernova_2009, chernova_2007, pmlr-v78-maeda17a} predicts a confidence score based on the current state of the agent, and requests for assistance when the confidence score does not meet a pre-defined threshold.
Our method is similar to these approaches in that we use a cosine distance metric to measure similarity from the semantic information present in the language descriptions of skills without any strict labels.


\label{sec:related_work}

\section{Conclusion}

In conclusion, we present a novel framework for robot agents to learn task relevant knowledge and skills from interactions with human users. To the best of our knowledge this is the first work to demonstrate end-to-end dynamic visuo-motor skill learning while querying a user with dialog to express doubt.  
Our agent can actively interact with human users and adapt its known skills to novel tasks by maintaining its semantic understanding of the tasks.
We outperform the existing continual learning baseline of GMM-LoRA in two separate simulated continual learning domains. 
Furthermore, our framework is able to learn a completely new visual-motor skill from human users (at $100\%$) with only a few robot demonstrations, without affecting the performance of any existing skills (at $91.67\%$)fulfilling continual learning requirements in robotics. 
Finally, we conducted a human-subjects experiment to demonstrate our framework's ability to complete tasks such as sandwich making from interactions with participants at a $81.25\%$ task success rate while demonstrating that after learning our participants spend a higher ratio of time solving their auxiliary distraction task than helping the robot solve the cooking tasks ($p< 0.001$). We believe such scalable interactive continual robot learning approaches would help non-experts work with robots more conveniently.
\newpage

\label{sec:conclusion}


\section{Limitations}
We present an approach to teach skills to robots using techniques from active learning and continual learning while using language as a modality to query and reason over the skills known to the agent.
As with any research this is a process of continual improvement. With that in mind here are our limitations for each part of our methodology.

\subsection{Limited Dialog Interactions.}
Firstly,  the dialog interactions are limited. 
Although LLMs allow more freedom in the dialog interactions than generating dialog with templates, the robot agent can only ask  questions or requests for help for the unknown skills in this sandwich making domain.
Secondly, our robot agent treats each skill action or dialog action in the fashion of an option framework. 
More specifically, our robot agent can only take a new action after the previous skill action or dialog action is completed, and cannot switch to a different action during the execution of an action. This makes the interactions rigid.
Thirdly, the turn-taking in our framework is tightly controlled, and not dynamic. Dynamic turn-taking happens in natural interactions between humans. This is an active open problem in the HRI community.
  
\subsection{Study Limitations.}
 We acknowledge that we need to conduct a wider user study with a larger number of skills and cooking tasks using our approach. 
 Our human-subjects study does not establish the efficiency of the different policy learning algorithms; this comparison was only done in the simulation experiments where we will demonstrate the efficacy of ACT-LoRA. Partly this was also because of the stability of existing algorithms, but maybe with more data from users this issue could have been fixed.

\subsection{Algorithmic Design Space Limitations.}
Our preliminary experiments on robot learning policies were conducted with limited robot data.
Models with larger visual backbones, such as TAIL, might achieve a better performance when provided with unlimited robot data which is currently scarce especially for our domain.
Furthermore, although the UNet diffusion policy with LoRA did not work well in our preliminary experiments, the capability of diffusion transformer~\cite{peebles2023scalablediffusionmodelstransformers} with LoRA remains unexplored in this work. 
We suspect that the LoRA can work well with diffusion transformers as the LoRA network was initially designed as an augmentation for the attention layers~\cite{hu2021lora}.
Further investigations in how the continual learning framework fits into different training algorithms and architectures for robotics is still left to be done.

Our ACT-LoRA approach while being sample efficient has been observed to have issues with heterogeneous demonstrations. Moreover, we want to scale to more contact rich tasks as seen in cooking but there are hardware limitations to tasks like cutting. During chopping and cutting the robot has to continue functioning after each collision of a chopping action where the blade meets the board. We removed such dynamic tasks from our study because the robot's collision model would not allow it to continue even though the formalism is capable of learning these behaviors. 

\subsection{Limited Generalizability.}
Although our continual learning policy has better generalizability than DMP or waypoint-based learning methods~\cite{grannen2024vocalsandboxcontinuallearning}, our COLADA agent is designed to work specifically on our sandwich making domain, and will not directly function on a completely different domain.
COLADA requires a different set of prompts to function on a completely different domain. These prompts need to include domain-specific examples that convert natural language commands into task tokens. 
We describe more details with examples on how to implement COLADA for a different domain in Appendix~\ref{app:recreation}.
\label{sec:limitations}


\newpage

\clearpage
\acknowledgments{If a paper is accepted, the final camera-ready version will (and probably should) include acknowledgments. All acknowledgments go at the end of the paper, including thanks to reviewers who gave useful comments, to colleagues who contributed to the ideas, and to funding agencies and corporate sponsors that provided financial support.}


\bibliography{example}  
\clearpage
\appendix
\section*{Overview of the Appendix}
\label{app:overview}
In the appendices below, Additional details regarding implementation details of the robot agents in the human-subjects study are provided, including details for both hardware configurations, other components of the agents such as speech recognition and text-to-speech, and the prompts we designed for the LLMs.
We also provide and discuss additional results on both the simulation experiments and human-subjects study along with the results of statistic tests.
Lastly, we will close the appendices with extended discussion on additional research opportunities and remaining open challenges that are relevant to this work.

This work is heavily grounded on our human-subjects study; video demonstrations of our study can be found on the anonymous associated webpage: \href{https://sites.google.com/view/corl-25-dialog/home}{https://sites.google.com/view/corl-25-dialog/home}

An overview of each appendix is as follows:

\noindent\rule{\linewidth}{0.5pt}

\hyperref[app:recreation]{\autoref{app:recreation}: \textit{Implementing COLADA}}

We provide implementation details for the COLADA agent in our human-subjects experiment, including hardware configurations, other components of the agents, prompts for LLMs, and discussion on how to re-implement COLADA on a different domain.

\begin{itemize}[leftmargin=3mm]
    \setlength\itemsep{5pt}
    \item[] \hyperref[app_sub:hardware]{\autoref{app_sub:hardware}: \textit{Hardware Configuration}} \\
    Details on the hardware configurations and other settings for the human-subjects study.
    \item[] \hyperref[app_sub:system_arch]{\autoref{app_sub:system_arch}: \textit{System Architecture}} \\
    Additional system details for speech recognition and text-to-speech components.
    \item[] \hyperref[app_sub:policy_implementation]{\autoref{app_sub:policy_implementation}: \textit{Details on Continual Learning Policies}} \\ 
    Details on implementations of our continual learning policies.
    \item[] \hyperref[app_sub:prompt]{\autoref{app_sub:prompt}: \textit{Details on LLM Interaction Module}} \\
    Prompting details on GPT-4.0 Turbo.
    \item[] \hyperref[app_sub:different_domain]{\autoref{app_sub:different_domain}: \textit{Implementing COLADA for a Different Domain}} \\
    Descriptions on what it takes to re-create COLADA for a different domain.
\end{itemize}

\bigskip

\hyperref[app:detailed_study]{\autoref{app:detailed_study}: \textit{Details of the Human-Subjects Study}}

We provide details on the pilot study and the procedures of our human-subjects study, additional experiment results, details for statistic tests. We also provide extended discussion on these experiment results and the limitations of our study.

\begin{itemize}[leftmargin=3mm]
    \setlength\itemsep{5pt}
    \item[] \hyperref[app_sub:detailed_procedure]{\autoref{app_sub:detailed_procedure}: \textit{Detailed Procedures}} \\
    Details on the procedures of our human-subjects study.
    \item[] \hyperref[app_sub:detailed_study_results]{\autoref{app_sub:detailed_study_results}: \textit{Additional Results and Statistic Tests}} \\
    Additional results and analysis of the human-subjects study, and the details of the statistic tests.
    \item[] \hyperref[app_sub:extended_discussion]{\autoref{app_sub:extended_discussion}: \textit{Additional Findings and Analysis}} \\
    Additional findings and analysis on the results of the human-subjects study.
    \item[] \hyperref[app_sub:extended_limitation]{\autoref{app_sub:extended_limitation}: \textit{Limitations of the Study}} \\
    Extended discussion on the limitation of the study design and potential improvements.
\end{itemize}

\bigskip
\hyperref[app:detailed_simulation]{\autoref{app:detailed_simulation}: \textit{Details of the Simulation Experiments}}

We provide details on the simulation experiments, including additional experiment results and extended analysis on model performance.

\bigskip

\newpage{}
\section{Implementing COLADA}
\label{app:recreation}
Implementing the COLADA on a real robot for a user study requires more than the learning components such as the LLM interaction module and the continual learning policy.
Hardware integrations are also needed for the robot to perform the tasks during the human-subjects study.
Furthermore, we also need to integrate other components such as real time speech recognition and text-to-speech to facilitate the dialog interactions between the users and the robot. 
In this section, we describe how we integrated these additional components in our system. We then provide additional details for our LLM interaction module, and provide guidelines on how to implement our COLADA agent on a different domain.

\subsection{Hardware Configuration}
\label{app_sub:hardware}

Our robotic setup includes a Franka FR3 Robot and three Realsense D435 cameras. 
We set up our cameras to provide a frontal view, a top-down view, and a wrist-mounted camera for a view from the robot's perspective.
This configuration allows us to capture dense and diverse features for training our policy. 
Our data collection pipeline includes a 6D Spacemouse from 3DConnexion, which dictates the motion of the robot end effector. This facilitates the collection of dense data. Although limited by the data collection rate, this setup allows users to control the robot in the task space with relative ease because of the intuitive nature of the Spacemouse.

Our workspace for the human-subjects study includes a table with items curated for the system. 
We designed 3D-printed tools tailored to support our task requirements as an attachment for the Franka Robot. 
These tools include a knife for cutting task and a spatula for spreading task, and are picked and placed using pre-specified waypoints.

\subsection{System Architecture}
\label{app_sub:system_arch}

Instead of asking users to provide textual inputs through keyboard, we use spoken language as the channel of communication between users and robots. This is because that the distribution of spoken language and textual inputs can be different, and spoken language is a more natural form of interactions for applications like household robots.
We use Whisper~\cite{radford2022robustspeechrecognitionlargescale} from OpenAI API endpoint to transcribe users' utterances into text inputs for the LLM. 
Off-the-shelf real time text-to-speech model from OpenAI TTS API is used to enable our robot agents to perform vocal dialog interactions with the human users.
The turns-takings in the dialog interactions are tightly controlled. We inform the human users when the robot agents are taking their utterances as inputs.

\subsection{Details on Continual Learning Policies}
\label{app_sub:policy_implementation}

\textbf{Implementation details for ACT-LoRA.} 
We describe the details of our implementation of the ACT-LoRA policy. Following \citet{zhao2023learning}, we train with a CVAE architecture and discard the additional encoder during inference. 
We adjust the number of parameters for different experiments accordingly. 
For all of our experiments, we use a $4$-layer transformer encoder both the CVAE encoder and the state encoder.
For the RLBench experiments and the real-world experiments, we use a hidden dimension of $2048$ and attention layers with $6$ heads.
For the LIBERO experiment, we use a hidden dimension of $256$ and attention layers with $8$ heads.
We extract features from raw image inputs from multiple cameras using resnet-$18$. These visual features are fed to the transformer encoder along with the proprioceptive inputs.
For the decoder side, we use $6$-layer transformer decoder for the real-world experiments and the RLBench experiments, and $4$-layer transformer decoder for the LIBERO experiments.
Trainable embeddings are used for all experiments. 
We also use a chunk size of $100$ as it gives the best performance empirically~\cite{zhao2023learning}. The same configuration is also used for the baseline ACT model.
As for the configuration of the low-rank adaptors, we follow TAIL~\cite{liu2024tail} and use a rank size of $8$ for all experiments. 
For both the simulation experiments and the human subject study, each skill is associated with a set of unique adaptor weights. 

\textbf{Implementation details for GMM-LoRA.} 
We re-implemented GMM-LoRA with the help from the authors of TAIL~\cite{liu2024tail} and the reference to the transformer-GMM policy from the LIBERO paper~\cite{liu2023liberobenchmarkingknowledgetransfer}. To reduce the computation cost for the original TAIL model, we use a transformer encoder in replacement to the GPT-2 temporal decoder. We also replace the CLIP image encoder with a resnet-18 model. For a fair comparison, we adjust the scale of the GMM-LoRA model to be similar to that of the ACT-LoRA model for each experiment.
The GMM-LoRA model takes in linguistic task descriptions, image observations, and proprioceptive inputs over history timesteps.
We first extract the feature of the raw image inputs and the linguistic task descriptions using the resnet-18 vision backbone and a frozen BERT text encoder. Then, we use a FiLM layer to inject the linguistic features into the image features and the proprioceptive inputs. These inputs are treated as the input tokens of the transformer temporal encoder. Then, we use an MLP layer to project the encoded tokens into parameters for Gaussian Mixture Models(GMM). During training, the model is optimized by minimizing the negative log-likelihood loss of the ground truth actions over multiple time-steps. During inference, we sample only one action from the distribution of the GMM predicted by the model.
Following TAIL~\cite{liu2024tail}, our GMM-LoRA model predicts an action chunk of size $10$. 
For fair comparison, we use a GMM-LoRA of similar scale to that of the ACT-LoRA.
For the LIBERO experiments, we use $8$-layer of transformer encoder with $6$ heads, with a hidden dimension of $256$. For the RLBench experiments, we use $10$-layer of transformer encoder with $8$ heads, with a hidden dimension of $2048$. We use a rank size of $8$ for all the adaptor weights introduced by the low-rank weights.

\textbf{Observations on GMM-based architecture.} In our simulation experiments and pilot studies, we observed that GMM-based architecture appears to struggle with joint-position based control.
This limitation is not studied in previous work that used GMM-based architecture~\citep{liu2023liberobenchmarkingknowledgetransfer, liu2024tail}.
We tested GMM-LoRA with joint-position control in the RLBench environment, and with operational space control(OSC) in the LIBERO environment.
As demonstrated in Table~\ref{tab:libero},\ref{tab:rlbench}, while GMM-LoRA achieves a reasonable performance in LIBERO that is comparable to the GMM transformer policy~\cite{liu2023liberobenchmarkingknowledgetransfer}, it struggles in all metrics in the RLBench environment.
Additionally, GMM-LoRA was unstable in our pilot study on the sandwich-making domain, where joint-position control is used.
It crashed the robot into the table twice, the predicted trajectories by GMM-LoRA for the real robot were not smooth.
Therefore, we suspect that GMM-LoRA model struggles with joint-position controls, and further studies are needed to verify this hypothesis.


\begin{figure*}
    \centering
    \includegraphics[width=1\textwidth]{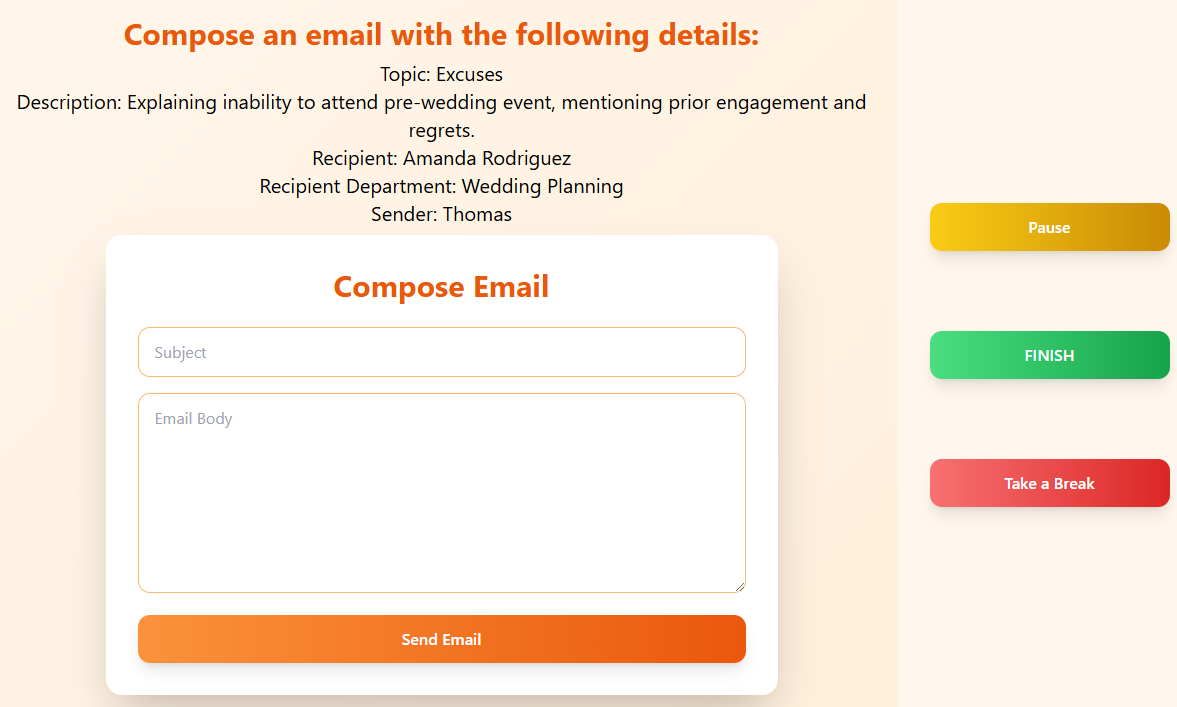}
    \caption{A screen shot of our email writing interface, where users compose emails with synthetic topics and recipient, such as excuses, device maintenance, inquiries to local attractions, etc. These synthetic email information are generated with ChatGPT and does not include any real information.}
    \label{fig:interface}
\end{figure*}

\subsection{Details on LLM Interaction Module}
\label{app_sub:prompt}
We use GPT-4.0 Turbo with function calling as the base for our LLM interaction module.
As described in the main paper, the interaction module has two major functionalities. 
We describe our prompting strategies at a high level, and the complete prompts will be released as a part of the code when the paper is accepted.
Firstly, we use the LLM to convert the initial dialog with the users into a sequence of skills. 
The LLM is prompted to continue the dialog interactions with users until it gets the confirmation from the users that the complete sequence of skills has been described.
This dialog history is then fed back into the LLM and converted into a sequence of task tokens, which will be used to compare semantic similarity.
Here, we prompt several examples of parsing initial dialog into skill tokens. We also use chain-of-thoughts prompting~\cite{wei2023chainofthoughtpromptingelicitsreasoning} using a style inspired by previous work~\cite{dai2024think}. These examples are from simulated scenarios of making sandwiches that are not used in our human-subjects study.
Secondly, when the COLADA agent and the inverse semantics agent encounter an unknown skill, we use the LLM to start dialog interactions with human users. 
For both the COLADA agent and the inverse-semantics agent, the LLM is prompted with the skill unknown to the robot agents to ask the human users to perform the corresponding skill.
After that, the COLADA agent is additionally prompted to request for robot demonstrations from the human users to learn the unknown skill.
Similarly, we provide examples of these interactions in the prompt, and use chain-of-thoughts prompting~\cite{wei2023chainofthoughtpromptingelicitsreasoning}.

\subsection{Implementing COLADA for a Different Domain}
\label{app_sub:different_domain}

With all the implementation details, we will close this section with a discussion on what it takes to implement the COLADA agent in a completely different domain, for example, a machine-shop assistant robot.

\textbf{Domain-specific dialog state machine.} Firstly, the COLADA agent needs a dialog state machine that is designed specifically for the domain because dialog interactions can be different and more complicated in a different domain.
In our sandwich-making domain, the COLADA agent only asks for task specification in the initial dialog interaction, and the other dialog interactions are limited to asking for help with unknown skills.
However, the dialog interactions can be different and more complicated for a different domain, as multiple rounds of specifications and clarifications might be needed, even if the robot possesses the skill to perform a task.
Consider the case where the COLADA agent is used as a machine-shop assistant robot, and the customer requires the robot to cut metal into certain shape.
In this scenario, the robot agent also needs to use dialog to ask for specifications and clarifications for the task, such as querying the desired shape and size from the customer. 
To handle these different dialog interactions, the COLADA agent needs a dialog state machine that is designed for the domain.
More specifically, the agent needs to have the knowledge of what information is missing to execute a task, and how to ask for that information via dialog interactions. This knowledge is domain-specific and needs to be incorporated into the dialog state machine.
In our example of the machine-shop assistant robot, the LLM needs to be prompted to understand specifications, and query corresponding clarification questions such as``What shape do you want to cut the metal into?'' or ``What is the color of the material?'' when specifications are not sufficient.

\textbf{Pre-train policy with domain-specific skills.} Secondly, the ACT-LoRA policy needs to be pre-trained on some common tasks in the domain. 
The ACT-LoRA policy can achieve its best performance if the pre-trained skills and the novel skills share the same domain. The weights of the base architecture have much more parameters($ \sim98\%$) than the weights of the task-specific adapters ($\sim 2\%$)~\cite{liu2024tail}. 
Pre-training the policy with skills from the same domain can greatly boost the performance of the policy on the novel skills as this exploits the domain knowledge stored in the base architecture weights learned from pre-training in the same domain. 
Continuing with the example of the machine-shop assistant robot, the robot should be pre-trained with necessary commonly-used basic skills, such as cutting metals and plastics, fastening screws, and drilling holes on woods.
With the basic commonly used skills in this domain learned, the policy can then continually learn other skills with few demonstrations such as installing specific attachments or nailing.

\textbf{Design necessary tools for robots.} Lastly, tools to support the corresponding tasks might be needed as the existing tools are designed for humans and the robot might not be able to use them. 
For the sandwich making domain, we customized a blade and a butter brush for the robot because our robot does not have the dexterity to pick up a knife and use it.
In the case of a machine-shop assistant robot, one might need to design specific tools such as hammers, glue guns, and screw drivers for robots, as the most common parallel gripper does not allow the robot to used these tools in a similar fashion to humans.


\begin{table*}[t]
    \centering
    \begin{adjustbox}{width=\linewidth}
    \begin{tabular}{cccccc}
        \toprule
        Agent    & Interruption Count & Normalized Completed Email Count & Normalized Word Count & Total Time & Task Time\\ 
        \midrule
        & \multicolumn{5}{c}{Phase One}\\
        \cmidrule(lr){2-6}
        COLADA& $2.13 \pm 0.13$ & $0.27 \pm 0.03$ & $0.24 \pm 0.01$ & $2176.67 \pm 57.06$ & $1035.21 \pm 26.10$\\
        Inverse Semantics& $1.13 \pm 0.09$ & $0.16 \pm 0.02$ & $0.20 \pm 0.01$ & $943.93 \pm 32.41$ & $753.21 \pm 25.85$\\
        Inarticulate& $0.00 \pm 0.00$ & $0.07 \pm 0.02$ & $0.08 \pm 0.01$ & $493.01 \pm 58.62$ & $412.98 \pm 56.69$ \\
        \midrule
        & \multicolumn{5}{c}{Phase Two}\\
        \cmidrule(lr){2-6}
        COLADA& $0.00 \pm 0.00$& $0.25 \pm 0.03$ & $0.23 \pm 0.02$ & $1083.42 \pm 27.28$ & $1033.70 \pm 26.32$\\
        Inverse Semantics& $1.00 \pm 0.00$ & $0.17 \pm 0.02$ & $0.17 \pm 0.01$ & $870.77 \pm 26.26$ & $738.27 \pm 24.02$\\
        Inarticulate& $0.00 \pm 0.00$ & $0.08 \pm 0.01$ & $0.07 \pm 0.01$ & $426.94 \pm 51.85$ & $376.78 \pm 48.74$\\
        \bottomrule
    \end{tabular}
    \end{adjustbox}
        \caption{The objective metrics of the human users on the distraction tasks of the study. The interruption count measures how many times each agent interrupt the users during the entire evaluation phase. The normalized email completion count measures the number of emails completed by the users while the agent is performing the task, normalized by the total number of emails completed by each user. The normalized word count measures the total number of words the users input when the agent is executing the tasks, normalized by the total number of words of each user for all agents. Total time measure the total amount of execution time in seconds of each agent, including the time that the agent interacts with the users and the time that the agent perform skills autonomously. Task time measures the amount of time in seconds for users to complete the distraction task, which is also the time that the agent performs skills autonomously.}\label{tab:distraction_metrics}
\end{table*}

\begin{table*}[]
\begin{adjustbox}{width=\linewidth}
    \begin{tabular}{ccccccc}
    \toprule
    
      Metrics   & SUS($\uparrow$)& Anthropomorphism($\uparrow$)  & Likability($\uparrow$)  &  Animacy($\uparrow$) &Perceived Intelligence($\uparrow$) & Comparative($\uparrow$)\\ 
      \midrule
        & \multicolumn{6}{c}{Phase One}\\
      \cmidrule(lr){2-7}
      COLADA &  $8.06\pm 1.61$ & $14.75\pm 0.89$ & $20.38\pm 0.77$ &$19.06\pm 0.85$ & $36.13\pm 0.92$& N/A \\
      Inverse Semantics &  $11.13\pm 1.38$ & $16.38\pm 0.98$ &$20.13\pm 0.69$ &$21.31 \pm 0.98$ & $37.31 \pm 0.89$ & N/A \\
      Inarticulate &  $4.06 \pm 2.64$ & $12.25 \pm 1.05$& $17.13 \pm 1.17$& $16.00 \pm 1.34$ & $29.31 \pm 1.72$ & N/A\\
      \midrule
        & \multicolumn{6}{c}{Phase Two}\\
      \cmidrule(lr){2-7}
      COLADA &  $12.50\pm 2.49$ & $15.94 \pm 1.04$ & $20.19 \pm 1.19$ &$20.63\pm 1.32$ & $35.69 \pm 1.67$ & $0.44 \pm 0.87$ \\
      Inverse Semantics &$9.31 \pm 2.55$   & $15.31 \pm 1.20$ &$19.94 \pm 1.07$ &$20.75 \pm 1.16$ & $36.00 \pm 1.47$ & $-0.63 \pm 0.68$ \\
      Inarticulate & $1.19 \pm 2.62$  & $12.00 \pm 0.94$ & $17.25 \pm 1.27$& $15.50 \pm 1.34$ & $29.25 \pm 1.95$ &$-5.81 \pm 0.86$\\
     
    \bottomrule
    \end{tabular}
\end{adjustbox}
\caption{The subjective metrics for the interaction phase. We use the same ACT-LoRA policy as the policy for all the three agents.}

\label{tab:subjective}
\end{table*}

\begin{table}[t]
        \centering
            \begin{tabular}{c|cc}
                \toprule
                Agent    & Interface Time Ratio(Phase one)       & Interface Time Ratio(Phase two) \\ 
                \midrule
                COLADA& $47.78 \pm 1.19$& $\mathbf{95.41 \pm 0.38}$ \\
                Inverse Semantics& $80.27 \pm 2.06$ & $85.13 \pm 2.46$ \\
                Inarticulate& $80.88 \pm 2.27$ & $86.82 \pm 1.46$  \\
                \bottomrule
            \end{tabular}%
        \caption{The ratio of the interface time for participants. This metric measures how many percent of the time the users spend on the distraction task of writing emails.}\label{tab:interface_time_ratio}

\end{table}

\section{Details of the Human-Subjects Study}
\label{app:detailed_study}

We describe the details of the human-subjects study. 
Our human-subjects study is approved by the Institutional Review Board(IRB) of the university.
We tested the study with $20$ pilots before conducting the experiments on the participants. We fixed the issues of unclear instructions, short execution times for the learned skills and ambiguous phrases when the LLM was asking questions. We had to fine-tune the prompts of the LLMs a lot so the robot asked questions pertinant to the task of sandwich making.
We also adjusted the configurations for the sandwiches, because some tasks can be very difficult for the novice users to teach the robot, such as picking up a deformable object.
Additionally, we made the interface of the distraction email writing task more intuitive for the participants, and created an instructional video for the email writing interface.
Our email writing interface is demonstrated as Fig.~\ref{fig:interface}.
All the instructional materials we used for the study can be found in the supplemental materials and the associated webpage.

\textbf{Challenges with GMM-LoRA on real robot.} We did not use GMM-LoRA model as a baseline policy in our human-subjects study for two major reasons. 
Firstly, we use joint-position control in the real-world sandwich making domain that is used for our human-subjects study, and GMM-LoRA appears to struggle in joint-position based control in our simulation experiments.
We evaluated GMM-LoRA on two simulated environments that use different controls for the robot:- RLBench that uses joint-position control and LIBERO that uses operational space control.
As shown in Table~\ref{tab:libero},\ref{tab:rlbench}, GMM-LoRA achieves a reasonable performance in the LIBERO environment but struggles in all metrics in the RLBench environment.
We therefore hypothesize that GMM-LoRA model struggles with joint-position controls. However, further study is needed to verify this hypothesis.
Secondly, we tested GMM-LoRA in our pilot studies, and does not have a stable performance to be used as a baseline in a human-subjects study. 
More specifically, the policy crashed the robot into the table when it tried to pick up a bowl.
We then examined GMM-LoRA's performance on other tasks in the real-world domain, and we observed that the policy tends to move the robot at a high speed, and the trajectories predicted by the policy were not smooth.
As a result, we consider the GMM-LoRA model is unsafe to be used in a human-subjects study.

For the actual study a total of $16$ participants were recruited through campus advertisements.
The study is composed of two separate phases, the interaction phase that takes $120$ minutes and the evaluation phase that takes $60$ minutes, with a voluntary participation. The participants, including the pilots, are compensated with $\$35$ Amazon gift card for their time.
We designed the two-phase study for two major reasons. Firstly, our COLADA agent requires five hours to train for the novel skill. Secondly, we want to demonstrate a thorough comparison for the workload and objective metrics on the distraction task between our COLADA agent and the inverse semantics agent in the two phases.
COLADA requires the users to remotely control the robot arm to perform the task in the interaction phase, and is fully automated in the evaluation phase, whereas the inverse semantics agent behaves the same in both phases by requesting the users to directly perform the task that it does not know.

\subsection{Detailed Procedure}
\label{app_sub:detailed_procedure}
We describe the detailed procedure for the study as follows.

\textbf{Interaction Phase.} Participants first filled out the consent form and a pre-study survey. Then, we handed out a general introduction of the experiment. The participants were then asked to read the instructions for the interaction phase, and watch a demonstration video. The demonstration video introduces how the robot agent requests for different types of help differently, and how to answer different requests from the robot agent. We use a completely different domain(Placing a block in the box) as example in the demonstration video. The instruction introduces domain relevant information, such as the configuration of the robot's workspace, the sandwich to make, and the steps to make the sandwich.
The participants then watch another demonstration video that introduces how to use the email writing interface.
The anonymized instructions and videos can be found in the supplementary material, and Fig.~\ref{fig:interface} shows our email writing interface. Then, the participants interacted with the three agents, the inarticulate agent, the inverse semantics agent, and the COLADA agent, in a random order. The inarticulate agent never interacts with the users except for getting the initial instruction set from the user. The inverse semantics agent always asks the human users for help when it encounters any task that it is uncertain with. The COLADA agent interacts with the human users by asking task-relevant question, asking for human help, and asking for robot demonstrations. 
The users then work on the distraction email writing tasks while these robot agents make the sandwich, and provide the required help from the agent when needed.
After interacting with each system, the participants were asked to fill-out a post-survey, including questions from NASA-TLX~\cite{tlx}, SUS~\cite{sus}, and $4$ sub-scales from the GodSpeed Questionnaire Series~\cite{godspeed}(Likability, Animacy, Natural, Perceived Intelligence). After the participants finished the interaction phase, we fine-tuned the ACT-LoRA policy the robot demonstrations collected from the users for COLADA.

\textbf{Evaluation Phase.} Participants came back to the lab. We handed the same instructions to the participants for them to ask the robot to make the same sandwich. The participants interacted with the same three robot agents, the inarticulate agent,the inverse semantics agent, and the COLADA agent. All the three agents remember the instructions to make the sandwich provided by the participants from the interaction phase. The inverse semantics agent and the inarticulate did not learn from the robot demonstrations from the interaction phase. This means that the inverse semantics agent still asked for help from the users for the same skill, and the inarticulate agent still failed to perform the same skill. The COLADA learned the novel skill from the demonstration in the interaction phase, and did not interact with the human users except for the initial interactions. After watching each agent, the participants were asked to fill out the same post-survey for the system. After watching all the three systems, the participants were asked to rank the three systems on $7$ different description(helpful, useful, efficient, competent, uncooperative, inefficient, incompetent).

\subsection{Additional Results and Statistic Tests}
\label{app_sub:detailed_study_results}

The objective results on the task completion and skill success rates are presented in Table~\ref{tab:study}, and the objective results on distraction tasks are presented in Table~\ref{tab:interface_time_ratio},\ref{tab:distraction_metrics}. We also present results on subjective metrics for both phases in Table~\ref{tab:subjective}.

Based on our analysis, we found that COLADA is more efficient in time for our participants in phase two than in phase one. Additionally, COLADA is more time efficient for the user than the inverse semantics agent in phase two.
For subjective metrics, no significance was found for the workload metrics between any agent pair. Both agents that can ask intelligent questions(COLADA and the inverse semantics agent) are considered better than the inarticulate agent in the sub-scales of system usability, anthropomorphism, likeability, animacy, perceived intelligence, and the comparative survey. Additionally, COLADA is considered better than the inverse semantics agent in the system usability sub-scale.
We perform a normality test with Shapiro-Wilk test for each metric. If the data from such metric passes the normality test($p>0.05$), we apply a parametric statistical test. Otherwise, we report the results of a non-parametric statistical test. 
The detailed results are described as follows.

\textbf{Users' ratio of time on distraction task.} Results from Shapiro-Wilk test suggest that conditions for normality were met for the data points to run a parametric statistical test($p=0.18, W=0.92$). Hence, we compare the time ratio metric between phase one and phase two for COLADA using paired t-test. Results from paired t-test suggest that COLADA allows users to spend more of their time on the email writing distraction task in phase two than in phase one($p<0.001, t=38.69$). 

Results from Shapiro-Wilk test suggest that conditions for normality were not met for the data points to run a parametric statistical test ($p=0.005, W=0.82$). Hence, we compare the time ratio metric between COLADA and the inverse semantics agent using Wilcoxon Signed-Rank test. Results from Wilcoxon Signed-Rank test suggest that user can spend more time on the email writing task working with COLADA than the inverse semantics agent in phase two($p<0.001, Z=4.17$).

\textbf{SUS.} Results from Shapiro-Wilk test suggest that conditions for normality were met for the data points to run a parametric statistical test($p=0.49, W=0.96$). Hence, we conduct a paired t-test to compare the system usability metric of COLADA with the inarticulate agent. Results from paired t-test suggest that COLADA is considered better than the inarticulate agent in the system usability sub-scale in phase two($p=0.002, t=1.83$).

Results from Shapiro-Wilk test suggest that conditions for normality were met for the data points to run a parametric statistical test($p=0.07, W=0.90$). Hence, we conduct a paired t-test to compare the system usability metric of the inverse semantics agent with the inarticulate agent. Results from paired t-test suggest that the inverse semantics agent is considered better than the inarticulate agent in the system usability sub-scale in phase two($p=0.006, t=2.82$).

Results from Shapiro-Wilk test suggest that conditions for normality were met for the data points to run a parametric statistical test($p=0.49, W=0.95$). Hence, we conduct a paired t-test to compare the system usability metric of COLADA with the inverse semantics agent. Results from paired t-test suggest that COLADA is considered better than the inverse semantics agent in the system usability sub-scale in phase two($p=0.04, t=1.83$).

\textbf{Anthropomorphism.} 
Results from Shapiro-Wilk test suggest that conditions for normality were met for the data points to run a parametric statistical test($p=0.34, W=0.94$). Hence, we conduct a paired t-test to compare the anthropomorphism metric of COLADA with the inarticulate agent. Results from paired t-test suggest that COLADA is considered better than the inarticulate agent in the anthropomorphism metric in phase two($p=0.003, t=3.18$).

Results from Shapiro-Wilk test suggest that conditions for normality were met for the data points to run a parametric statistical test($p=0.78, W=0.97$). Hence, we conduct a paired t-test to compare the anthropomorphism metric of the inverse semantics agent with the inarticulate agent. Results from paired t-test suggest that the inverse semantics agent is considered better than the inarticulate agent in the anthropomorphism metric in phase two($p=0.01, t=2.54$).

\textbf{Likability.} 
Results from Shapiro-Wilk test suggest that conditions for normality were met for the data points to run a parametric statistical test($p=0.57, W=0.95$). Hence, we conduct a paired t-test to compare the likeability metric of COLADA with the inarticulate agent. Results from paired t-test suggest that COLADA is considered better than the inarticulate agent in the likeability metric in phase two(($p=0.04, t=1.86$).

Results from Shapiro-Wilk test suggest that conditions for normality were met for the data points to run a parametric statistical test($p=0.59, W=0.96$). Hence, we conduct a paired t-test to compare the likeability metric of the inverse semantics agent with the inarticulate agent. Results from paired t-test suggest that the inverse semantics agent is considered better than the inarticulate agent in the likeability metric in phase two($p=0.03, t=2.00$).

\textbf{Animacy.} Results from Shapiro-Wilk test suggest that our data in the animacy metric does not satisfy the condition for a parametric test($p=0.03, W=0.87$). Hence, we conduct a Wilcoxon Signed-Rank test to compare the animacy metric of COLADA with the inarticulate agent. Results from the Wilcoxon Signed-Rank test suggest that COLADA is considered better than inarticulate agent by users in the animacy metric with significance in phase two($p<0.001, t=3.10$). 

Results from Shapiro-Wilk test suggest that our data in the animacy metric satisfies the condition for a parametric test($p=0.07, W=0.90$). Results from paired t-test suggest that the inverse semantics agent is considered better than inarticulate agent by users in the animacy metric with significance in phase two($p<0.001, t=3.87$).

\textbf{Perceived Intelligence.} Results from Shapiro-Wilk test suggest that conditions for normality were met for the data points to run a parametric statistical test($p=0.07, W=0.90$). Hence, we conduct a paired t-test to compare the perceived intelligence metric of COLADA with the inarticulate agent. Results from paired t-test suggest that COLADA is considered better than the inarticulate agent in the perceived intelligence metric in phase two($p=0.01, t=2.58$).

Results from Shapiro-Wilk test suggest that conditions for normality were met for the data points to run a parametric statistical test($p=0.12, W=0.91$). Hence, we conduct a paired t-test to compare the perceived intelligence metric of the inverse semantics agent with the inarticulate agent. Results from paired t-test suggest that the inverse semantics agent is considered better than the inarticulate agent in the perceived intelligence metric in phase two($p=0.003, t=3.09$).

\textbf{Comparative.} Conditions for normality were not met for the data points to run a parametric statistical test($p=0.028, W=0.871$). 
Hence, we conducted a Wilcoxon Signed-Rank test to compare COLADA with the inarticulate agent in the comparative metric. Results from Wilcoxon Signed-Rank test suggest that COLADA is preferred by user in the direct comparison with the inarticulate agent with significance($p=0.004, Z=2.61$).

Conditions for normality were met for the data points to run a parametric statistical test($p=0.24, W=0.93$). Hence, we applied a paired t-test to compare the inverse semantics agent against the inarticulate agent in the comparative metric. Results from Wilcoxon Signed-Rank test suggest that the inverse semantics agent is preferred by user in the direct comparison with the inarticulate agent with significance($p=0.001, Z=3.02$).

\subsection{Additional Findings and Analysis}
\label{app_sub:extended_discussion}
We hypothesized that the users experience higher workload for COLADA than the inverse agent in the interaction phase, and a lower workload for the COLADA than the inverse semantics in the evaluation phase because we consider that for remotely controlling the robot arm to complete the task requires higher workload than directly completing the task themselves for the users, and the fully automated robot agent requests the least workload.
We reject our hypothesis and accept the null hypothesis of --  there is no difference in the users's perception of workload between COLADA and the inverse semantics agent.
We consider that the workload from the distraction email writing task can be the major confounding factor to the workload metrics. 
From the users' perspective, even though COLADA saves their time by finishing the sandwich autonomously, they still need to work longer on the distraction tasks as the robot takes longer time to finish the same task than taking the users' help. 
As a result, the users might not perceive that the fully automated COLADA agent invokes less workload than the inverse semantics agent, and we did not find any significance in the subjective workload metric. 
However, our objective metrics that measure the ratio of time that users spend on the distraction tasks indicate that our COLADA agent allows user to use more of their time on the distraction time than the inverse semantics agent in phase two($p<0.001, Z=3.61$). This shows that a fully automated learning agent is more efficient for the users. Additionally, we observed that COLADA achieves a higher ratings than the inverse semantics agent with significance($p=0.04, t=1.83$) in the System Usability Scales(SUS). This demonstrates that a learning system is considered more useful than a system that relies on humans' help by the users.

We hypothesized that COLADA allows users to spend less time interacting with the COLADA agent in the test phase than in the interaction phase. 
According to a paired t-test, we find that COLADA allows participants to spend more time on finishing the distraction email writing in the test phase than in the interaction phase with significance($p<0.001, t=38.69$).
This is because that the subjects need to spend more time teaching unknown skills to the COLADA agent in the interaction phase, and a learned COLADA agent in the test phase barely requires any time from the users. 
These results further suggest that the efficiency of a learning agent can be continually improved over time and experiences.

We also hypothesized that the inarticulate agent is considered worse by the users than the other agents that ask intelligent questions. Our results from Table~\ref{tab:subjective} suggest that our participants consider that both COLADA and the inverse semantics agent better than the inarticulate agent in SUS, the Likeability, Animacy, Perceived Intelligence, and Anthropomorphism sub-scales from the Godspeed Questionnaire Series, and our customized comparative survey with significance. This indicates that even knowing that a skill is unknown is sufficient to demonstrate intelligence and be more useful to a human user.

\subsection{Limitations of the Study}
\label{app_sub:extended_limitation}
There are two major limitations on the human-subjects study.
Firstly, we need to increase the scale of the study to better understand the robustness of COLADA and ACT-LoRA. Currently, limited by the scale of data, we only conducted the study with two different sandwich configurations on 8 different tasks. A scaled-up version of the study with more tasks, more data, and more users will be necessary to test the robustness of our framework.
Secondly, the demographic of the study is limited to university students. More subjects with wider demographic distribution will be needed to show that COLADA can work with the general population.
Lastly, our human-subjects study does not establish the efficiency of the different policy learning algorithms; this comparison was only done in the simulation experiments where we will demonstrate the efficacy of ACT-LoRA. Partly this was also because of the stability of existing algorithms, but maybe with more data from users this issue could have been fixed.

\section{Details of Simulation Experiments}
\label{app:detailed_simulation}

\subsection{Detailed Results on RLBench}
\label{app_sub:detailed_rlbench}
\begin{table*}[t]
\centering
\resizebox{\linewidth}{!}{%
\begin{tabular}{c|ccccccc}
\toprule
Model    & Pre-trained Skills(1000 traj.)   & Fine-tune Skills(1000 traj.) & Overall Success Rate(1000 traj.)& Fine-tune Skills(100 traj.) & Overall Success Rate(100 traj.) & Fine-tune Skills(5 traj.) & Overall Success Rate(5 traj.)\\ 
\midrule
ACT-LoRA& $\mathbf{60.75 \pm 2.40}$ & $54.00 \pm 9.73 ^ *$  & $\mathbf{59.40 \pm 1.52}$ & $47.67\pm 10.24^*$ & $\mathbf{58.87 \pm 1.55}$& $77.67 \pm 9.36$ & $\mathbf{64.13 \pm 1.80}$  \\
GMM-LoRA& $26.08 \pm 4.02$ &  $13.33 \pm 4.50$ & $23.53 \pm 2.99$ & $11.00 \pm 4.07$& $23.73 \pm 3.15$ & $16.67 \pm 4.92$& $24.20 \pm 3.72$\\
ACT& $9.25 \pm 2.51$ & $62.00 \pm 8.84 ^ *$   & $19.80 \pm 1.69$ & $63.33 \pm 9.90 ^*$ & $20.60 \pm 1.11$&   $\mathbf{95.00 \pm 4.22}$ & $26.40 \pm 2.45$ \\

\bottomrule
\end{tabular}%
}
\caption{Complete experimental results on RLBench dataset. $*$ indicates that two models have a similar best performance. ACT-LoRA out performs ACT and GMM-LoRA in the overall success rates and has fewer issues with forgetting pre-trained skills. GMM-LoRA is based on SOTA TAIL~\cite{liu2024tail} model with a smaller visual backbone which can be fine-tuned for a smaller set of tasks.}
\label{tab:complete_rlbench}
\end{table*}

\begin{table*}[t]
\centering
\resizebox{\linewidth}{!}{%
\begin{tabular}{c|ccccccccccccccc}
\toprule
Model    & close door & close fridge & meat off grill & meat on grill & open box & open door & open window & phone on base & put money in safe & put rubbish in bin & slide block to target & take lid off sauce pan & toilet seat down & turn tap & water plants\\ 
\midrule
& \multicolumn{15}{c}{Pre-trained(1000 traj.)}\\
\cmidrule(lr){2-16}
ACT-LoRA& $1.25 \pm 1.25$ & $96.25\pm 2.39$ & $72.50 \pm 24.28$ & $71.25 \pm 22.11$ & $63.75 \pm 22.49$ & $78.75 \pm 16.38$ & $73.75 \pm 24.61$ & $58.75 \pm 19.83$ & $61.25 \pm 20.65$ & $52.50 \pm 17.85$ & $46.25 \pm 16.63$ & $71.25 \pm 23.84$ & $93.75 \pm 6.25$ & $45.00 \pm 15.41$ & $25.00 \pm 7.36$ \\
GMM-LoRA& $1.25 \pm 1.25$ & $80.00 \pm 10.61$ & $6.25 \pm 3.15$ & $12.50 \pm 7.77$ & $37.50 \pm 15.34$ & $36.25 \pm 9.66$ & $41.25 \pm 16.63$ & $3.75 \pm 3.75$ & $28.75 \pm 13.29$ & $1.25 \pm 1.25$ & $0.00 \pm 0.00$ & $28.75 \pm 12.81$ & $70.00 \pm 7.36$ & $16.25 \pm 7.18$ & $27.50 \pm 7.77$ \\
ACT& $0.00 \pm 0.00$ & $72.50 \pm 14.22$ & $0.00 \pm 0.00$ & $0.00 \pm 0.00$ & $0.00 \pm 0.00$ & $1.25 \pm 1.25$ & $0.00 \pm 0.00$ & $1.25 \pm 1.25$ & $0.00\pm 0.00$ & $0.00\pm 0.00$ & $0.00\pm 0.00$ & $0.00\pm 0.00$ & $50.00 \pm 13.39$ & $12.50 \pm 9.46$ & $1.25 \pm 1.25$  \\
\midrule
& \multicolumn{15}{c}{Fine-tuned(1000 Traj.)}\\
\cmidrule(lr){2-16}
ACT-LoRA& $5.00$ & $90.00$ & $75.00$ & $90.00$ & $45.00$ & $80.00$ & $65.00$ & $25.00$  & $65.00$ & $10.00$ & $5.00$ & $95.00$ & $85.00$ & $25.00$ & $50.00$ \\
GMM-LoRA& $0.00$ & $70.00$ & $0.00$ & $15.00$ & $0.00$ & $25.00$ & $0.00$ & $5.00$ & $0.00$ & $0.00$ & $0.00$ & $0.00$ & $65.00$ & $20.00$ & $0.00$ \\
ACT& $5.00$ & $95.00$ & $90.00$ & $90.00$ & $65.00$ & $85.00$ & $15.00$ & $80.00$ & $45.00$ & $85.00$ & $15.00$ & $90.00$ & $100.00$ & $50.00$ & $20.00$  \\
\midrule
& \multicolumn{15}{c}{Fine-tuned(100 Traj.)}\\
\cmidrule(lr){2-16}
ACT-LoRA& $0.00$ & $100.00$ & $85.00$ & $75.00$ & $55.00$ & $85.00$ & $30.00$ & $15.00$ & $50.00$ & $5.00$ & $5.00$ & $90.00$ & $100.00$ & $20.00$ & $0.00$  \\
GMM-LoRA& $0.00$ & $80.00$ & $0.00$ & $5.00$ & $5.00$ & $20.00$ & $0.00$ & $0.00$ &$0.00$ &$0.00$ &$0.00$ &$0.00$ & $25.00$ & $25.00$ & $5.00$\\
ACT& $15.00$ & $95.00$ & $90.00$ & $75.00$ & $65.00$ & $65.00$ & $60.00$ & $70.00$ & $65.00$ & $55.00$ & $20.00$ & $90.00$ & $100.00$ & $55.00$ & $30.00$  \\
\midrule
& \multicolumn{15}{c}{Fine-tuned(5 Traj., Static evaluation)}\\
\cmidrule(lr){2-16}
ACT-LoRA& $0.00$ & $100.00$ &$100.00$ & $100.00$& $100.00$&$100.00$ &$100.00$ &$100.00$ & $0.00$ & $60.00$ &$100.00$ & $100.00$& $100.00$&$100.00$ & $5.00$ \\
GMM-LoRA& $0.00$ & $0.00$ &  $0.00$&  $0.00$&  $0.00$& $15.00$ & $0.00$ & $0.00$ & $0.00$ & $5.00$ & $35.00$ & $25.00$ & $85.00$ & $80.00$ & $5.00$ \\
ACT&$100.00$ &$100.00$ &$100.00$ &$100.00$ &$100.00$ &$100.00$ &$100.00$ &$100.00$ & $35.00$ &$100.00$ &$100.00$ &$100.00$ &$100.00$ &$100.00$ &$90.00$  \\
\bottomrule
\end{tabular}%
}
\caption{Experimental results on each skill of the RLBench dataset. We report success rate of each skill under pre-trained and fine-tuned with different number of trajectories. As we perform a five-fold validation on the skills, the statistics of the pre-trained skills come from $4$ models, whereas the success rates of the fine-tuned skills come from the evaluation of a single model. For each model, we evaluate each skill by rolling out the skill in the simulator for $20$ times.}
\label{tab:rlbench_per_skill}
\end{table*}

We present the complete experimental results of the three policies in the RLBench simulator. 
We perform five-fold validation on $15$ selected tasks from the RLBench simulator, and present the results in Table~\ref{tab:complete_rlbench}.
Detailed performance of each skill is presented in Table~\ref{tab:rlbench_per_skill}.
Column \textbf{Pre-trained skills($x$ traj.)} measures the policies' average success rate on the skills that policies are pre-trained on after fine-tuning on $x$ demonstration trajectories. Columns \textbf{Fine-tuned skills($x$ traj.)} and \textbf{Overall Success Rate($x$ traj.)} measure the policies' average success rate on the new skills, and the average success rate for both the pre-trained and fine-tuned skills respectively. 

All the three models are trained to predict joint positions in RLBench, and went through the same pre-trained, fine-tuned training schema.
During the pre-train phase, each model is trained with $1000$ robot demonstrations from each pre-train task for $5$ epochs. 
In the fine-tuning phase, we only train the weights introduced by the Low-Rank Adaptor for ACT-LoRA and GMM-LoRA, while the ACT model is fine-tuned with all its weights.
We fine-tuned models for $10, 100$ and $1000$ epochs when using $1000, 100$ and $5$ trajectories for fine-tune skills respectively.
Notice that due to the limitation of the visual-motor policies, we use a static location to evaluate the fine-tune tasks when we fine-tune with $5$ robot trajectories for all models. For the pre-trained skills and fine-tuned skills trained with more trajectories, we use a randomized initial configuration in evaluation. 

As shown in Table~\ref{tab:complete_rlbench} and Table~\ref{tab:rlbench_per_skill}, the full fine-tuned ACT model achieves a strong performance on fine-tuned skills, demonstrating its strong capability of learning fine-grained control. However, it suffers a near zero success rate for most of the pre-trained skills after fine-tuning. This shows that ACT suffers from catastrophic forgetting and can no longer perform the pre-train tasks after fine-tuning. 
On the contrary, our ACT-LoRA model not only achieves a comparable performance on fine-tuned skills as the ACT model, but also outperforms other baselines in pre-trained skills and overall success rate. This demonstrates that our ACT-LoRA model can continually learn novel skills without suffering from catastrophic forgetting.

GMM-LoRA model performs the worst in both pre-trained skills and fine-tune skills on RLBench dataset. This is to our surprise as GMM-based model has demonstrated a strong performance in controlling robot manipulators on LIBERO dataset~\citep{liu2023liberobenchmarkingknowledgetransfer, liu2024tail}. We suspect that the reason for the poor performance is that GMM-based model suffers from joint-position controls, but further investigations are needed to verify this hypothesis.


\subsection{Detailed Results on LIBERO}
\label{app_sub:detailed_libero}
\begin{table*}[t]
\centering
\resizebox{\linewidth}{!}{%
\begin{tabular}{c|cccccccccc}
\toprule
Model    & 0 & 1   & 2 & 3 & 4 & 5 & 6 & 7 & 8 & 9 \\ 
\midrule
& \multicolumn{9}{c}{Pre-trained(50 Traj.)}\\
\cmidrule(lr){2-11}
ACT-LoRA& $70.00 \pm 7.07$ & $48.75 \pm 18.53$ & $80.00 \pm 3.54$ & $63.75 \pm 21.35$& $57.50 \pm 10.90$& $65.00 \pm 21.89$& $95.00 \pm 2.04$& $65.00 \pm 22.27$& $47.50 \pm 6.61$& $61.25 \pm 4.73$ \\
GMM-LoRA& $72.50 \pm 10.90$& $38.75 \pm 10.08$ & $95.00 \pm 2.04$ & $60.00 \pm 20.00$ & $53.75 \pm 5.54$ & $36.25 \pm 15.86$& $82.50 \pm 2.50$& $63.75 \pm 21.93$& $75.00 \pm 2.04$ & $70.00 \pm 4.08$ \\
ACT& $0.00 \pm 0.00$ & $0.00 \pm 0.00$ & $1.25 \pm 1.25$ & $1.25 \pm 1.25$ & $0.00 \pm 0.00$ & $0.00 \pm 0.00$ & $0.00 \pm 0.00$ & $0.00 \pm 0.00$ & $0.00 \pm 0.00$ & $0.00 \pm 0.00$\\
\midrule
& \multicolumn{9}{c}{Fine-tuned(50 Traj.)}\\
\cmidrule(lr){2-11}
ACT-LoRA& $40.00$ & $65.00$ & $45.00$ & $65.00$ & $15.00$ & $65.00$ & $40.00$ & $25.00$ & $25.00$ & $20.00$ \\
GMM-LoRA& $0.00$ & $0.00$&$0.00$&$50.00$&$0.00$& $5.00$& $0.00$& $35.00$& $0.00$& $0.00$ \\
ACT& $65.00$ & $45.00$& $80.00$& $90.00$& $55.00$& $75.00$& $85.00$& $80.00$& $35.00$& $75.00$ \\
\midrule
& \multicolumn{9}{c}{Fine-tuned(5 Traj.)}\\
\cmidrule(lr){2-11}
ACT-LoRA& $35.00$ & $85.00$& $5.00$& $55.00$& $10.00$& $45.00$& $60.00$& $35.00$& $5.00$& $20.00$ \\
GMM-LoRA& $0.00$ & $10.00$& $0.00$& $30.00$& $0.00$& $20.00$& $0.00$&$0.00$&$0.00$ & $0.00$ \\
ACT& $60.00$& $85.00$& $70.00$& $75.00$& $40.00$& $65.00$& $45.00$& $40.00$& $30.00$& $40.00$ \\
\bottomrule
\end{tabular}%
}
\caption{Experimental results on each skill of LIBERO-spatial dataset. We report success rate of each skill under pre-trained and fine-tuned with different number of trajectories. As we perform a five-fold validation on the skills, the statistics of the pre-trained skills come from $4$ models, whereas the success rates of the fine-tuned skills come from the evaluation of a single model. For each model, we evaluate each skill by rolling out the skill in the simulator for $20$ times.}
\label{tab:libero_spatial}
\end{table*}

\begin{table*}[t]
\centering
\resizebox{\linewidth}{!}{%
\begin{tabular}{c|cccccccccc}
\toprule
Model    & 0 &1   & 2 & 3& 4 & 5 & 6 & 7 & 8 & 9 \\ 
\midrule
& \multicolumn{9}{c}{Pre-trained(50 Traj.)}\\
\cmidrule(lr){2-11}
ACT-LoRA& $85.00 \pm 4.56$& $37.50\pm 13.62$ & $87.50 \pm 6.61$& $37.50 \pm 13.62$& $86.25\pm 4.27$& $31.25 \pm 13.90$& $92.50\pm 3.23$& $65.00 \pm 22.08$&  $82.50 \pm 7.22$ & $65.00 \pm 11.37$  \\
GMM-LoRA& $93.75 \pm 3.15$ & $63.75 \pm 17.84$& $96.25 \pm 1.25$& $61.25 \pm 20.55$& $88.75 \pm 5.15$& $62.50 \pm 21.07$& $88.75 \pm 5.54$& $52.50 \pm 14.79$& $87.50 \pm 5.20$& $82.50 \pm 7.77$ \\
ACT& $2.50 \pm 2.50$ & $0.00 \pm 0.00$& $1.25\pm 1.25$& $0.00 \pm 0.00$& $5.00 \pm 3.54$& $13.75 \pm 13.75$& $37.50 \pm 21.65$& $7.50 \pm 4.33$& $47.50 \pm 27.50$& $13.75 \pm 9.44$ \\
\midrule
& \multicolumn{9}{c}{Fine-tuned(50 Traj.)}\\
\cmidrule(lr){2-11}
ACT-LoRA& $75.00$ & $25.00$& $65.00$& $35.00$& $70.00$& $65.00$& $100.00$& $90.00$& $100.00$&  $55.00$\\
GMM-LoRA& $0.00$ & $50.00$ & $0.00$ & $5.00$ & $0.00$ & $45.00$ & $0.00$ & $50.00$& $0.00$& $0.00$ \\
ACT&  $50.00$& $25.00$& $90.00$& $35.00$& $70.00$& $40.00$& $95.00$& $95.00$& $60.00$& $70.00$ \\
\midrule
& \multicolumn{9}{c}{Fine-tuned(5 Traj.)}\\
\cmidrule(lr){2-11}
ACT-LoRA& $10.00$& $80.00$& $45.00$& $25.00$& $30.00$& $55.00$& $95.00$& $80.00$& $60.00$& $0.00$ \\
GMM-LoRA& $0.00$ & $15.00$ & $0.00$ & $65.00$& $0.00$& $15.00$& $0.00$& $45.00$& $0.00$& $0.00$ \\
ACT& $75.00$& $25.00$& $60.00$& $55.00$& $15.00$& $15.00$& $35.00$& $25.00$& $45.00$& $5.00$ \\
\bottomrule
\end{tabular}%
}
\caption{Experimental results on each skill of LIBERO-object dataset. We report success rate of each skill under pre-trained and fine-tuned with different number of trajectories. As we perform a five-fold validation on the skills, the statistics of the pre-trained skills come from $4$ models, whereas the success rates of the fine-tuned skills come from the evaluation of a single model. For each model, we evaluate each skill by rolling out the skill in the simulator for $20$ times.}
\label{tab:libero_object}
\end{table*}

\begin{table*}[t]
\centering
\resizebox{\linewidth}{!}{%
\begin{tabular}{c|cccccccccc}
\toprule
Model    & 0 &1   & 2 & 3& 4 & 5 & 6 & 7 & 8 & 9 \\ 
\midrule
& \multicolumn{9}{c}{Pre-trained(50 Traj.)}\\
\cmidrule(lr){2-11}
ACT-LoRA& $78.75 \pm 4.73$& $72.50 \pm 24.28$& $91.25\pm 3.75$& $31.25\pm 11.25$& $90.00 \pm 4.08$& $63.75 \pm 21.93$& $78.75 \pm 3.15$& $70.00 \pm 23.80$& $78.75\pm 5.15$& $81.25 \pm 6.25$ \\
GMM-LoRA& $92.50 \pm 3.23$& $71.25 \pm 22.21$& $98.75 \pm 1.25$& $26.25 \pm 8.75$& $93.75\pm 1.25$& $61.25\pm 21.45$& $72.50 \pm 1.44$& $72.50\pm 20.97$& $82.50 \pm 6.29$& $82.50\pm 4.33$ \\
ACT& $0.00 \pm 0.00$&$0.00 \pm 0.00$&$0.00 \pm 0.00$&$0.00 \pm 0.00$&$0.00 \pm 0.00$&$0.00 \pm 0.00$&$0.00 \pm 0.00$&$0.00 \pm 0.00$&$0.00 \pm 0.00$&$0.00 \pm 0.00$ \\
\midrule
& \multicolumn{9}{c}{Fine-tuned(50 Traj.)}\\
\cmidrule(lr){2-11}
ACT-LoRA& $65.00$& $95.00$& $55.00$& $25.00$& $85.00$& $0.00$& $60.00$& $0.00$& $45.00$& $60.00$ \\
GMM-LoRA& $0.00$& $40.00$& $0.00$& $0.00$& $0.00$& $10.00$& $0.00$& $55.00$& $0.00$& $0.00$ \\
ACT& $15.00$& $15.00$& $15.00$& $35.00$& $45.00$& $10.00$& $40.00$& $5.00$& $15.00$ & $0.00$\\
\midrule
& \multicolumn{9}{c}{Fine-tuned(5 Traj.)}\\
\cmidrule(lr){2-11}
ACT-LoRA& $10.00$&$90.00$& $15.00$& $0.00$& $60.00$& $10.00$& $0.00$& $10.00$& $30.00$& $5.00$ \\
GMM-LoRA& $0.00$ & $30.00$& $0.00$& $0.00$&$0.00$& $5.00$&$0.00$&$0.00$&$0.00$&$0.00$ \\
ACT&$0.00$ & $20.00$& $5.00$&$0.00$&$5.00$&$0.00$&$5.00$& $50.00$& $20.00$& $0.00$ \\
\bottomrule
\end{tabular}%
}
\caption{Experimental results on each skill of LIBERO-goal dataset. We report success rate of each skill under pre-trained and fine-tuned with different number of trajectories. As we perform a five-fold validation on the skills, the statistics of the pre-trained skills come from $4$ models, whereas the success rates of the fine-tuned skills come from the evaluation of a single model. For each model, we evaluate each skill by rolling out the skill in the simulator for $20$ times.}
\label{tab:libero_goal}
\end{table*}

We present the major results on the three task suites of the LIBERO dataset in Table~\ref{tab:libero}.
Additionally, we present the detailed performance of each skill from three suites of the LIBERO dataset in Table~\ref{tab:libero_spatial},\ref{tab:libero_object},\ref{tab:libero_goal}. 
Column \textbf{Pre-trained skills($x$ traj.)} measures the policies' average success rate on the skills that policies are pre-trained on after fine-tuning on $x$ demonstration trajectories. Columns \textbf{Fine-tuned skills($x$ traj.)} and \textbf{Overall Success Rate($x$ traj.)} measure the policies' average success rate on the new skills, and the average success rate for both the pre-trained and fine-tuned skills respectively. 

For each of the three suite of the LIBERO dataset, we apply the same training schema and perform a five-fold validation on the $10$ tasks of the task suite. 
All the three models are trained with robot trajectories in the operational space control(OSC), and went through the same pre-trained, fine-tuned training schema.
During the pre-train phase, each model is trained with $50$ robot demonstrations from each pre-train task for $100$ epochs. 
In the fine-tuning phase, we only train the weights introduced by the Low-Rank Adaptor for ACT-LoRA and GMM-LoRA, while the ACT model is fine-tuned with all its weights.
To study the models' performance with different data scales, we fine-tuned models for $1000$ and $100$ epochs when using $5$ and $50$ trajectories for fine-tune skills respectively.

As shown in Table~\ref{tab:libero}, we can observe that ACT-LoRA achieves the most stable performance across the three policies. In overall success rate, ACT-LoRA is either comparable to or better than a strong GMM-LoRA baseline. Additionally, although GMM-LoRA achieves the best performance in pre-trained skills in all the three task suites, ACT-LoRA outperforms GMM-LoRA on fine-tuned skills under all configurations without compromising much in the performance on the pre-trained skills. This demonstrates that ACT-LoRA is more suitable for continual learning than GMM-LoRA. On the other hand, ACT-LoRA shares the best performance in majority metrics on fine-tuned skills with an ACT model that undergoes full fine-tuning. However, ACT-LoRA achieves a significantly better performance than ACT in pre-trained skills and overall success rate metrics across all the three task suites. 
This demonstrates that ACT-LoRA is the most stable policy for continual learning when compared to the other strong baselines.

\end{document}